\documentclass[10pt,twocolumn,letterpaper]{article}

\usepackage{cvpr}              

\usepackage{pgfplots}
\usepackage{graphicx}
\usepackage{amsmath}
\usepackage{multirow} 
\usepackage{amssymb}
\usepackage{booktabs}
\usepackage{color, colortbl}
\usepackage{xspace}
\usepackage{adjustbox}
\usepackage{tablefootnote}
\usepackage{xcolor}
\usepackage[accsupp]{axessibility} 
\definecolor{LightCyan}{rgb}{0.88,1,1}
\definecolor{grey}{rgb}{0.9,0.9,0.9}

\definecolor{cvprblue}{rgb}{0.21,0.49,0.74}
\usepackage[pagebackref,breaklinks,colorlinks,citecolor=cvprblue]{hyperref}
\usepackage{tcolorbox}
\usepackage{xcolor}
\usepackage{dblfloatfix}
\usepackage{pgfplots}
\usepackage{algorithm}
\usepackage{amsmath}
\usepackage{algpseudocode}
\usepackage{pgfplotstable}
\usepackage{filecontents}
\usepackage{mdframed}
\pgfplotsset{compat=1.17}
\usetikzlibrary{backgrounds,datavisualization.formats.functions}

\pgfplotsset{compat=1.17}
\def\paperID{8276} 
\def\confName{CVPR}
\def\confYear{2025}

\title{\textsc{OSLoPrompt}: Bridging Low-Supervision Challenges and Open-Set Domain Generalization in CLIP}

\author{
Mohamad Hassan N C$^{1}$ \ Divyam Gupta$^{1}$ \ Mainak Singha$^{1}$ \ Sai Bhargav Rongali$^{1}$ Ankit Jha$^{2}$ \and Muhammad Haris Khan$^{3}$ \ Biplab Banerjee$^{1}$ \\  
{\normalsize $^{1}$Indian Institute of Technology Bombay} \quad  
{\normalsize $^{2}$The LNM Institute of Information Technology (LNMIIT)} \quad \\  
{\normalsize $^{3}$Mohamed Bin Zayed University of Artificial Intelligence}  
}

\begin{document}
\maketitle
 \begin{abstract}

We introduce Low-Shot Open-Set Domain Generalization (LSOSDG), a novel paradigm unifying low-shot learning with open-set domain generalization (ODG). While prompt-based methods using models like CLIP have advanced DG, they falter in low-data regimes (e.g., 1-shot) and lack precision in detecting open-set samples with fine-grained semantics related to training classes. 
To address these challenges, we propose \textsc{OSLoPrompt}, an advanced prompt-learning framework for CLIP with two core innovations. First, to manage limited supervision across source domains and improve DG, we introduce a domain-agnostic prompt-learning mechanism that integrates adaptable domain-specific cues and visually guided semantic attributes through a novel cross-attention module, besides being supported by learnable domain- and class-generic visual prompts to enhance cross-modal adaptability. 
Second, to improve outlier rejection during inference, we classify unfamiliar samples as ``unknown'' and train specialized prompts with systematically synthesized pseudo-open samples that maintain fine-grained relationships to known classes, generated through a targeted query strategy with off-the-shelf foundation models. This strategy enhances feature learning, enabling our model to detect open samples with varied granularity more effectively. 
Extensive evaluations across five benchmarks demonstrate that \textsc{OSLoPrompt} establishes a new state-of-the-art in LSOSDG, significantly outperforming existing methods.\footnote{\url{https://github.com/has97/Osloprompt}}
\end{abstract}    
 \section{Introduction}

\begin{figure}
    \centering
    \includegraphics[width=0.9\columnwidth]{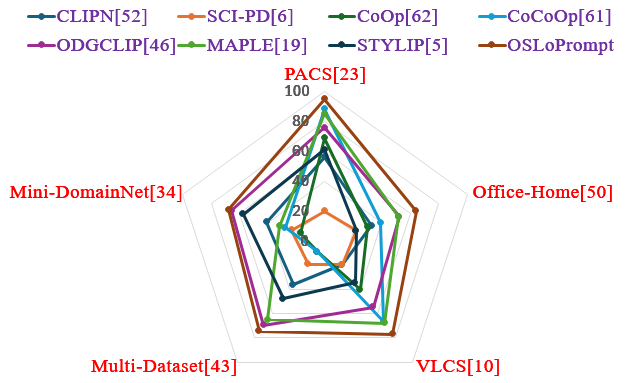}
    
    \vspace{-3mm}\caption{\textbf{Harmonic score (H-score) (between known and novel class performances) comparisons} of various CLIP-based DG/ODG/open-set recognition techniques versus our approach in LSOSDG setting with one-training example per known class, demonstrating the improved performances of \textsc{OSLoPrompt}.}
    \vspace{-5mm}
    \label{fig:perf_comp}
\end{figure}

Domain Generalization (DG) \cite{dg1} enhances model robustness by training across multiple source domains to enable generalization to unseen targets. Traditional DG typically operates within a fully-supervised, closed-set framework, leveraging abundant labeled data and assuming semantic alignment across domains \cite{l2g, style-neophile, crossgrad}. However, fields like healthcare often face data scarcity, motivating Few-Shot DG (FSDG) \cite{2lm}, which extends conventional few-shot learning \cite{wang2020generalizing} in the DG setting. Yet, the closed-set assumption falls short in real-world applications, where models encounter novel classes dynamically. ODG \cite{daml, odg-net} addresses this by incorporating mechanisms to reject outliers during inference but assumes a fully supervised training regimen, overlooking low-shot scenarios.

To this end, we introduce LSOSDG, a novel paradigm extending conventional ODG by imposing data scarcity on training classes (1/5-shot). Unlike domain-adaptive few-shot open-set learning \cite{pal2023domain}, which relies on predefined meta-training and meta-testing domains, LSOSDG requires learning a domain-generic classifier from limited source supervision while detecting outliers at inference without prior knowledge. This setup is ideal for dynamic, open-world scenarios—such as autonomous vehicles adapting to shifting conditions or medical diagnostics identifying emerging diseases—where data is sparse, domain shifts are unpredictable, and new classes appear spontaneously.

\begin{figure}
    \centering
    \includegraphics[width=\columnwidth]{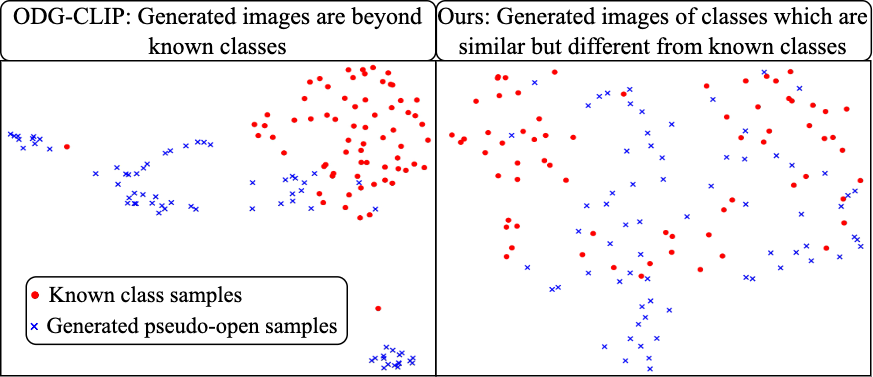}
    \vspace{-5mm}
    \caption{t-SNE \cite{tsne} of \textbf{known-class} and \textbf{pseudo-open samples} generated by ODG-CLIP \cite{odgclip} (left) and our method (right). Our approach produces fine-grained pseudo-open samples, creating a sharper closed-open class boundary and enhancing feature learning, resulting in an improvement over \cite{odgclip} on Mini-DomainNet \cite{domainnet} (Table \ref{tab:combined_ablation}), significantly boosting open-set detection.}
    \vspace{-5mm}
    \label{fig:visual}
\end{figure}

Existing CNN-based FSDG models \cite{2lm, urt} struggle in LSOSDG due to conflicting supervision requirements. In contrast, domain-agnostic prompt learning\footnote{Domain-agnostic prompts are designed in a way that can be applied to any unknown target domain without prior knowledge.} in VLMs like CLIP \cite{clip} shows promise for DG by visual content and style conditioned prompt learning \cite{stylip, applenet}. However, these methods are challenged in LSOSDG, where undefined open spaces and limited supervision hinder visual style discernment and performance (Fig. \ref{fig:perf_comp}).

ODG-CLIP \cite{odgclip} addresses ODG with a unified \texttt{Unknown} class prompt and diffusion-generated pseudo-open training samples for the outlier class. While effective in ODG, it falters in low-shot settings, akin to \cite{stylip}. Additionally, these pseudo-open samples lack the nuanced distinctions needed to separate visually coherent known and open-set samples, a notable gap in ODG research (Fig. \ref{fig:visual}).

Besides, our analysis (Table \ref{tab:combined_ablation}) highlights three main limitations in current domain-agnostic prompt modeling for LSOSDG. \textbf{First}, existing methods over-rely on textual cues, missing the synergy between visual and textual prompts needed for robust, domain-agnostic features in multi-domain DG. Effective DG requires learnable visual prompts to complement textual ones---a frequently overlooked aspect. \textbf{Second}, DG prompting approaches like \cite{odgclip, stylip} use learnable contexts \cite{coop} that lack the structured knowledge found in manually crafted prompts, as demonstrated by KG-CoOp \cite{kgcoop}; we argue for a hybrid prompt strategy to reduce misguidance under limited supervision. \textbf{Finally}, while recent work \cite{attr1, attr2, attr3, kim2024aapl, laclip, labo} shows that attribute-based prompting improves generalization through class-sharable attributes, these approaches often isolate visual and textual attributes. Instead, we advocate cross-referencing visual and textual attributes for richer semantic explanations of visual objects with a frozen CLIP backbone.

\noindent \textbf{Our contributions:} We propose a novel model, Open-Set LOw-shot PROMPT learning (\textsc{OSLoPrompt}), to address these challenges, with two key innovations:

\begin{figure}[t]
        \centering
        \vspace{-5mm}\includegraphics[width=\linewidth]{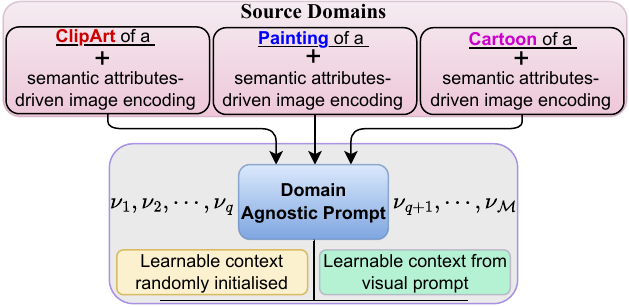}
        \vspace{-7mm}
        \caption{\textbf{Proposed prompt learning}: We develop a novel strategy for learning domain-agnostic prompts with tokens $\{\nu_{1:q:\mathcal{M}}\}$, inheriting context from source-specific prompts enriched with image-to attributes encodings. Some tokens also integrate knowledge from visual prompts spanning all training domains and classes. We differ considerably from the DG literature.}
        \label{fig:fig_teaser}
        \vspace{-5mm}
    \end{figure}

\textbf{- Fine-grained pseudo-open sample synthesis}: Building on \cite{odgclip}, we introduce a novel prompt to classify open samples as \texttt{Unknown} and employ pseudo-open image synthesis to train this prompt. Unlike the generic pseudo-open sample generation in \cite{odgclip}, which struggles with detecting fine-grained open samples, we propose a controlled synthesis approach where generated pseudo-open samples retain fine-grained semantic discrimination with respect to the training classes (Fig. \ref{fig:visual}). This is achieved by strategically prompting GPT-4o \cite{achiam2023gpt} and stable diffusion \cite{stablediffusion}.

\textbf{- Effective prompt learning in low-data regime}: In \textsc{OSLoPrompt}, we improve domain-agnostic semantic prompt learning by addressing two key challenges: enhancing prompt versatility across domains and categories, and bridging the visual-semantic gap effectively.

To make them universally applicable, we align the domain-agnostic prompt's context tokens with manually crafted source-domain-specific counterparts enriched with semantic attribute-based visual encodings. These attributes, generated by GPT-4o, capture class-specific details and are further refined through a novel image-to-attributes cross-attention module. This structured regularization enhances both domain independence and semantic richness.

To connect visual and semantic information further, we introduce a multi-modal prompting approach. Here, a learnable visual prompt within CLIP’s image encoder captures diverse source-domain features, which are then used to initialize some of the tokens in the agnostic prompts. This setup aligns the prompts more closely with the visual characteristics in the data, supporting better generalization (Fig. \ref{fig:fig_teaser}).
 In summary, our major contributions are:

\noindent \textbf{[-]} We introduce the LSOSDG problem setting, highlighting the limitations of existing DG methods and presenting a comprehensive solution through \textsc{OSLoPrompt}.

\noindent \textbf{[-]} We enhance prompt generalization using structured regularization, visually-driven semantic attribute guidance, and visual prompt learning. Additionally, we propose synthesizing superior pseudo-open samples to train a versatile outlier detection strategy through targeted prompting.

\noindent \textbf{[-]} We benchmark \textsc{OSLoPrompt} in LSOSDG setting on five datasets, backed by thorough ablation studies.
 \section{Related Works}
\textbf{DG and FSDG}: 
DG was introduced to address domain shifts by training models across diverse source domains, thereby enhancing performance on unseen targets \cite{dg1, l2g, style-neophile}. Beyond traditional methods, recent advancements in prompt-based approaches within VLMs \cite{huang2023sentence, clipood, stylip, odgclip, cheng2024disentangled} have shown considerable improvements in DG.
 In contrast, FSDG \cite{urt, tsa, cnaps, dgselect, 2lm} aims to generalize with limited supervision. It relies on transferrable knowledge from a distinct set of classes, often utilizing meta-learning and domain-specific adaptation. Nevertheless, we note that LSOSDG and FSDG follow distinct problem settings.

\noindent \textbf{ODG}: ODG extends DG to handle unknown classes during inference, a concept initially introduced by \cite{daml} using domain-augmented meta-learning. Unlike Open Set Recognition (OSR) \cite{osr1, osr2} and Open Set Domain Adaptation (OSDA) \cite{osda}, ODG faces the unique challenge of operating in an inductive setting without target domain data during training. MEDIC \cite{medic} advanced this field by matching domain and class-wise gradients, while methods like \cite{odg-net} focus on disentangled feature learning and adversarial sample synthesis for open-set detection. Prompt tuning in foundation models, such as those based on CLIP, and generalizable prompting techniques \cite{clipn, coop, cocoop, stylip} have also been adapted for ODG. However, these methods often struggle with diverse domains and establishing optimal confidence thresholds for detecting open samples. ODG-CLIP \cite{odgclip} introduces an \texttt{unknown} class for the outliers and generates pseudo-open samples using a diffusion model \cite{stablediffusion} to train the respective prompt. Concurrent approaches like \cite{scipd} explore style perturbation and knowledge distillation but generally speaking, all of them face difficulties with fine-grained differentiation between known and open classes.

\noindent \textbf{VLMs and prompt learning}: Vision-Language foundation models such as CLIP \cite{clip}, ALIGN \cite{align}, LiT \cite{lit}, FILIP \cite{filip}, and Florence \cite{florence} have made significant strides in image recognition by leveraging large-scale image-text pairings to capture rich multi-modal representations. Despite their strength in open-vocabulary tasks, adapting these models to specific challenges while preserving generalization remains complex. Approaches like CoOp \cite{coop}, CoCoOp \cite{cocoop}, MaPLe \cite{maple}, PromptSrc \cite{promptsrc}, and Kg-CoOp \cite{kgcoop} enhance token embeddings for task-specific adaptation, while recent methods such as \cite{attr1, attr2, attr3, kim2024aapl, laclip, labelme, desco} incorporate textual or visual attributes into prompts to better generalize from base to novel classes (more discussions in \textbf{Sup Mat}). Furthermore, \cite{hirohashi2024prompt} explores prompt regularization for low-shot training but is limited to typical zero-shot inference. Contrary to the literature, we introduce a comprehensive domain-agnostic and enhanced prompting strategy to handle the nuanced structure of LSOSDG, showing a more balanced performance on both the known and open-set classes than other counterparts (Table \ref{tab_open_1}).

 \section{Proposed Methodology}
\begin{figure*}
    \centering
    \includegraphics[width=0.9\linewidth]{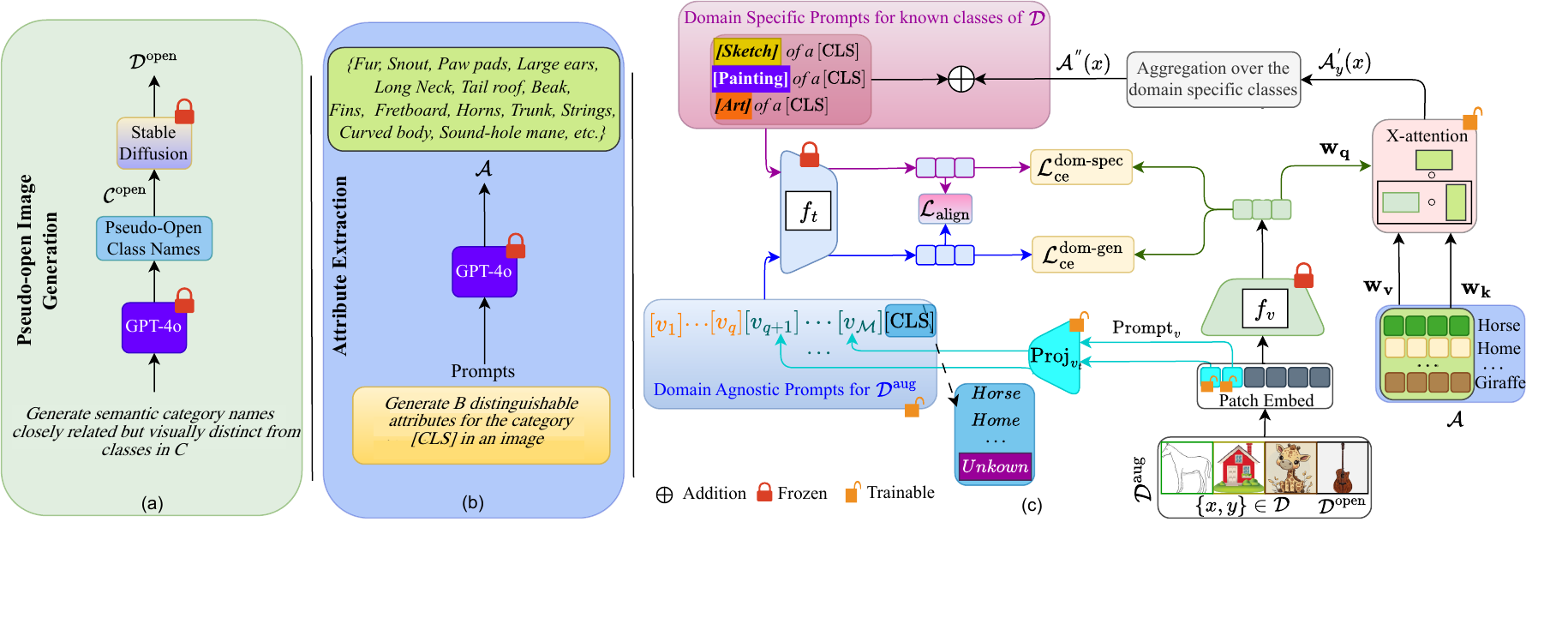}
    \vspace{-10mm}
    \caption{\textbf{Working principles of \textsc{OSLoPrompt}}. \textbf{(a)} Fine-grained pseudo-open samples $\mathcal{D}^{\text{open}}$ are generated using stable diffusion with pseudo-open class names $\mathcal{C}^{\text{open}}$ from GPT-4o. \textbf{(b)} GPT-4o generates attributes for each class in $\mathcal{C}$. \textbf{(c)} \textsc{OSLoPrompt} learns domain-agnostic prompts using tokens ${\nu_{1:\mathcal{M}}}$. The first $q$ tokens follow \cite{coop}, while tokens $q+1$ to $\mathcal{M}$ are initialized via learnable visual prompts, and transformed through a projector $\text{Proj}_{vt}$. Domain-agnostic prompts are regularized by domain-specific prompts enhanced with visually-guided semantic attributes, encoded through a cross-attention module with parameters ($\mathbf{w_k}, \mathbf{w_v}, \mathbf{w_q}$). The model is trained with a context alignment loss $\mathcal{L}_{\text{align}}$, along with visual-textual classification losses, handling known class samples for domain-specific prompts with $\mathcal{D}$ ($\mathcal{L}_{\text{ce}}^{\text{dom-spec}}$) and both known and pseudo-open class samples for domain-agnostic prompts with $\mathcal{D}^{\text{aug}}$ ($\mathcal{L}_{\text{ce}}^{\text{dom-gen}}$).}
    \label{fig:model}
    \vspace{-5mm}
\end{figure*}
We consider a scenario with $\mathcal{N}$ source domains, each defined as $\mathcal{D}_s = \{(x_i^s, y_i^s)\}_{i=1}^{n_s}$ where $x^s \in \mathcal{X}_s$ are input images and $y^s \in \mathcal{Y}_s$ are the corresponding labels. These domains, denoted by $\mathcal{D} = \{\mathcal{D}_s\}_{s=1}^{\mathcal{N}}$, have unique distributions ($\mathcal{P}(\mathcal{D}_s) \neq \mathcal{P}(\mathcal{D}_{s'})$ for $s \neq s'$) and feature a mix of shared and domain-specific classes, often resulting in class imbalance. Classes across the source domains are collectively represented by $\mathcal{C}=\bigcup_{s=1}^{\mathcal{N}} \mathcal{Y}_s$, with each class typically having limited samples (e.g., 1-shot or 5-shot).
During testing, the model encounters an unlabeled target domain $\mathcal{T}$ with its dataset $\mathcal{D}_t = \{x_j^t\}_{j=1}^{n_t}$, where $\mathcal{P}(\mathcal{D}_t)$ differs from $\mathcal{P}(\mathcal{D}_s)$ for all source domains. The target domain's label set, $\mathcal{Y}_t$, includes both known classes, $\mathcal{Y}_t^{\text{known}} = \mathcal{C}$, and novel classes or outliers, $\mathcal{Y}_t^{\text{novel}} = \mathcal{Y}_t \setminus \mathcal{Y}_t^{\text{known}}$.

To address the LSOSDG challenge, we frame it as a $|\mathcal{C}| + 1$-class classification task within the CLIP framework, as inspired by \cite{odgclip}. We design prompts for each of the $|\mathcal{C}|$ known classes and an additional \texttt{Unknown}-class prompt for cohesively classifying outlier instances during inference, without utilizing open-set knowledge during training.

Our approach, \textsc{OSLoPrompt}, focuses on two key strategies: \textbf{(a)} We synthesize robust pseudo-open samples that are proximal to known classes in the embedding space to train the \texttt{Unknown} class prompt. This ensures clear separation and reduces inference ambiguity. \textbf{(b)} We develop a domain-agnostic prompt learning strategy that distills contexts from structured, hand-crafted domain-specific prompts with semantic attribute encodings derived from the visual space, besides utilizing insights from learnable, domain- and class-generic visual prompts. The detailed architecture of \textsc{OSLoPrompt} is illustrated in Fig. \ref{fig:model}. Further specifics for both methods are provided below.

At the onset, we denote the frozen CLIP text and image encoders as $\mathcal{F}_t$ and $\mathcal{F}_v$, respectively. The textual encodings for the $s^{\text{th}}$ source domain and class label $y$ are represented as \textit{[$\text{Domain}_s$]} and \textit{[$\text{CLS}_y$]}.
\subsection{Synthesis of fine-grained pseudo-open images}
To generate fine-grained pseudo-open samples that are closely related to known classes, we leverage Stable Diffusion \cite{stablediffusion}, which offers substantial improvements over previous methods like \cite{l2g, odg-net, cumix} that used adversarial examples or manually combined image pairs to craft pseudo-open samples. By utilizing Stable Diffusion, our approach maintains domain-specific stylistic coherence and carefully controls semantic drift from known classes in \(\mathcal{C}\). This stylistic consistency is essential for accurately distinguishing between closed and open classes; unmanaged style variations can lead models to mistake domain shifts for class differences---a limitation seen in prior methods \cite{l2g, odg-net}.

Unlike \cite{odgclip}, which uses generic prompts in Stable Diffusion to generate out-of-support images as pseudo-open samples---often yielding semantically irrelevant images---our method enables controlled pseudo-open sample generation. The approach in \cite{odgclip} risks producing images that inadequately define decision boundaries, thus limiting the classifier's precision in segregating nuanced open-set samples from the known classes (Fig. \ref{fig:images_fine}).

To achieve this controlled generation, we first curate a set of pseudo-open class names, \(\mathcal{C}^{\text{open}}\), derived from \(\mathcal{C}\) (see \textbf{Sup Mat} for examples) to ensure similarity yet distinctiveness, guided by a targeted GPT-4o prompt:

\begin{mdframed}[backgroundcolor=gray!20, linewidth=0pt, leftmargin=1cm, rightmargin=1cm]
    ``\textit{Generate semantic category names closely related but visually distinct from classes in $\mathcal{C}$.}"
\end{mdframed}

For each class in \(\mathcal{C}^{\text{open}}\), we use Stable Diffusion to synthesize images in the chosen source-domain-specific style (\textit{Sketch, Clipart}, etc.), using prompts such as:

\begin{mdframed}[backgroundcolor=gray!20, linewidth=0pt, leftmargin=1cm, rightmargin=1cm]
    ``\textit{Generate images in the style of {[Domain]} depicting {[class from} \(\mathcal{C}^{\text{open}}\)\textbf{]}.}"
\end{mdframed}

$\mathcal{D}^{\text{open}}$ denotes the synthesized images in this way. Furthermore, we introduce the augmented source dataset, $\mathcal{D}^{\text{aug}}$ by combining $\mathcal{D}$ with $\mathcal{D}^{\text{open}}$.

\subsection{Proposed prompt learning strategy}
Our domain-agnostic semantic prompt-learning framework is structured around three core concepts:

\textbf{First}, we implement a multi-modal prompting approach that enhances CLIP’s capacity to learn visual abstractions invariant to domain and class distinctions, supporting robust DG. These learnable visual prompts also serve to initialize portions of the domain-agnostic prompts, improving feature encoding and reducing style biases. 
\textbf{Second}, we align the domain-agnostic prompt's contexts with those of the manually crafted, domain-specific prompts applied across all source domains in \(\mathcal{D}\). While these domain-specific prompts capture unique domain features, their adaptability is limited by static design. 
\textbf{Third}, to address this limitation, we enrich the domain-specific prompts with visually guided class attribute encodings. Using a cross-attention mechanism, each image is represented as a combination of class attributes generated by GPT-4o, enabling the prompts to capture detailed visual semantics and improve versatility across domains.
We detail these paradigms in the following.

\noindent \textbf{(i) Proposed generic visual prompting:} We incorporate the visual prompting strategy from \cite{vpt} within the visual encoder $\mathcal{F}_v$. For images in the augmented dataset $\mathcal{D}^{\text{aug}}$, additional visual prompts, denoted as \(\mathbf{{Prompt}_v} = \{p_v^{1:m}\}\) with $m$ tokens, are introduced at the first ViT layer of $\mathcal{F}_v$ (denoted as $\mathcal{F}_v^1$) alongside the initial image patch embeddings $(E_0)$. The forward pass through $\mathcal{F}_v$ is defined as:

\begin{equation}\centering
[c_1, \_ , E_1] = \mathcal{F}_v^1([c_0, \mathbf{{Prompt}_v}, E_0])
\end{equation}

\begin{equation}\centering
[c_l, E_l] = \mathcal{F}_v^l([c_{l-1}, E_{l-1}]), \quad l = 2, 3, \dots
\end{equation}

Here, \(c_l\) denotes the conventional learnable token embedding of ViT after the $l^{\text{th}}$ layer, and $[:, :]$ represents stacking and concatenation. The influence of the learned prompts $\mathbf{{Prompt}_v}$ is thereby integrated into any subsequent forward pass of $\mathcal{F}_v(x)$ during evaluation.

\noindent \textbf{(ii) A novel image-to-semantic attributes connection, improving the domain-specific prompts}: We begin by defining a general formulation for manually curated, domain-specific prompts for each source domain \( s \in [1, \mathcal{N}] \) and its respective classes \( y^s \in \mathcal{Y}_s \):

\begin{equation}
\mathbf{{Prompt}_s^{y^s}} = ``\textit{[$\text{Domain}_s$] of a [$\text{CLS}_{y^s}$]}"
\end{equation}

To enrich these prompts under limited supervision, we inject fine-grained, attribute-level knowledge into them; however, moving beyond static prompts with class descriptions \cite{attr1}, which often overlook the visual space, we seek to model the attribute's distributions within the images. This leads to our attribute-based image encoding scheme. Specifically, for each class \( y^s \), we define a set of \(\mathcal{B}\) domain-invariant attributes \( \mathcal{A}_{y^s} = [a_{y^s}^1, a_{y^s}^2, \dots, a_{y^s}^{\mathcal{B}}] \), generated via GPT-4o with a structured prompt:

\begin{mdframed}[backgroundcolor=gray!20, linewidth=0pt, leftmargin=1cm, rightmargin=1cm]
    ``\textit{Generate \(\mathcal{B}\) distinguishable attributes for the category [CLS] in an image.}"
\end{mdframed}

These attributes, consistent across domains, capture detailed semantic cues. For instance, generated attributes for \texttt{Dog} include \textit{Fur, Snout, Tail, Paw Pads} (Fig. \ref{fig:model}; additional details in \textbf{Sup Mat}).

To integrate these attributes into visual encoding, we use a cross-attention mechanism that enriches each image-label pair \( (x^s,y^s) \in \mathcal{D}_s \) with class-specific attributes \( \mathcal{A}_{y^s} \). Here, image features \( \mathcal{F}_v(x^s) \) act as the query, while semantic attributes \( \mathcal{F}_t(\mathcal{A}_{y^s}) \) serve as keys and values, with learnable projections \( \mathbf{w_q} \), \( \mathbf{w_k} \), and \( \mathbf{w_v} \) applied to obtain transformed representations $\mathcal{F}_v^q(), \mathcal{F}_t^k(), \mathcal{F}_t^v()$. The attribute-enhanced embedding for class \( y^s \) in image \( x^s \) is computed as:

\begin{equation}
\mathcal{A}'_{y^s}(x^s) = \operatorname{Softmax}\left[\frac{\mathcal{F}^q_v(x^s)\mathcal{F}^k_t(\mathcal{A}_{y^s})^T}{\sqrt{d}}\right] \mathcal{F}^v_t(\mathcal{A}_{y^s})\label{eq:ca}
\end{equation}

To obtain a class-agnostic encoding, we compute an average across classes:

\begin{equation}
\mathcal{A}''(x^s) = \frac{1}{|\mathcal{Y}_s|} \sum_{{y^{s'}} \in \mathcal{Y}_s} \mathcal{A}'_{y^{s'}}(x^s)
\end{equation}

The final domain-specific prompt combines the base prompt \( \mathbf{{Prompt}_s^{y^s}} \) with this averaged attribute embedding \( \mathcal{A}''(x^s) \) through token-wise addition:

\begin{equation}
\overline{\mathbf{{Prompt}_s^{y^s}}}(x^s) = \mathbf{{Prompt}_s^{y^s}} + \mathcal{A}''(x^s)
\end{equation}

The resulting prompt, \( \overline{\mathbf{{Prompt}_s^{y^s}}}(x^s) \), is dynamic and discriminative, capturing nuanced visual variations beyond the static base prompt \( \mathbf{{Prompt}_s^{y^s}} \) and enhancing domain-agnostic prompts substantially.

We train these prompts using a visual-textual contrastive objective:

{\small
\begin{equation}
\text{\footnotesize$
\mathcal{L}_{\text{ce}}^{\text{dom-spec}} = \operatorname*{min}_{\substack{\mathbf{w^q}, \mathbf{w^k} \hspace{0.2cm} \\ \mathbf{w^v}, \mathbf{Prompt}_v}} \sum_{s=1}^{\mathcal{N}} \mathbb{E}_{(x^s,y^s) \sim \mathcal{P}(\mathcal{D}_s)} \left[-\log p(y^s | x^s)\right]$
}
\label{eq:ce}
\end{equation}
}

where the probability \( p(y^s | x^s) \) is computed based on cosine similarity \( \delta \) and a temperature parameter \( \tau \):

{\small
\begin{equation}
\begin{aligned}
\text{\footnotesize $
p(y^s | x^s) = 
\frac{\exp\left(\delta\left(\mathcal{F}_t\left(\overline{\mathbf{Prompt}_s^{y^s}}(x^s)\right), \mathcal{F}_v(x^s)\right) / \tau\right)}
{\sum\limits_{y^{s'} \in \mathcal{Y}_s} \exp\left(\delta\left(\mathcal{F}_t\left(\overline{\mathbf{Prompt}_s^{y^{s'}}}(x^s)\right), \mathcal{F}_v(x^s)\right) / \tau\right)}
$}
\end{aligned}
\label{eq:compact}
\end{equation}
}

\begin{table*}[htbp]
\vspace*{-1.5mm}
\caption{Comparative analysis across five datasets in the 1-shot(top) and 5-shot (bottom) LSOSDG setting, reporting both the average closed-set accuracy (Acc) and H-score over all domain combinations, following a leave-one-domain-out protocol. Best performance in \textbf{bold} and the second-best in \textbf{Red}. Results on the domain combinations are mentioned in \textbf{Sup Mat}.}
\vspace*{-5mm}
\begin{center}
\scalebox{0.675}{
\begin{tabular}{lcccccccccccc|cc}
\toprule
\multicolumn{1}{l}{\multirow{2}{*}{\textbf{Methods}}} & \multicolumn{1}{c}{\multirow{2}{*}{\textbf{CLIP-based}}} & \multicolumn{1}{c}{\multirow{2}{*}{\textbf{Venue}}} 
& \multicolumn{2}{c}{\textbf{PACS}}& \multicolumn{2}{c}{\textbf{VLCS}} & \multicolumn{2}{c}{\textbf{OfficeHome}} & \multicolumn{2}{c}{\textbf{Multi-Dataset}} & \multicolumn{2}{c}{\textbf{Mini-DomainNet}} & \multicolumn{2}{|c}{\textbf{Average}} \\
\cmidrule(lr){4-5}\cmidrule(lr){6-7}\cmidrule(lr){8-9}\cmidrule(lr){10-11}\cmidrule(lr){12-13}\cmidrule(lr){14-15}
& & &\textbf{Acc} & \textbf{H-score} &\textbf{Acc} & \textbf{H-score} & \textbf{Acc} & \textbf{H-score} & \textbf{Acc} & \textbf{H-score} & \textbf{Acc} & \textbf{H-score} & \textbf{Acc} & \textbf{H-score} \\
\midrule

\cellcolor{cyan!10}CLIP + OpenMax (OSR) \cite{osr1} & \centering$\checkmark$ &CVPR'16& 20.24 & 31.97 & 20.59 & 31.83 & 20.00 & 32.64 & 11.74 & 20.87 & 16.92 & 28.05 & 17.90 & 29.07 \\

\cellcolor{cyan!10}CLIPN (OSR) \cite{clipn} & \centering $\checkmark$ & ICCV'23 &64.03 & 55.79 &25.34&19.49& 44.18 & 32.83 & 39.84 & 36.28 & 47.63 & 40.91 &44.20 &37.06 \\

\cellcolor{grey}MORGAN (FS-OSR) \cite{morgan} & \centering $\times$ & WACV'23& 37.40 & 19.06 &31.35&27.22& 19.21 & 18.51 & 30.00 & 37.26 & 22.40 & 15.70 &28.07&23.55\\

\cellcolor{cyan!10}\textsc{StyLIP} (DG + OSR)    \cite{stylip} & \centering $\checkmark$  &WACV'24& \cellcolor{red!20}74.89 & 60.99 &27.94 &34.61& \cellcolor{red!20}52.34 & 21.95 & 51.50 & 47.64 & 59.44 & 57.46 &53.22&44.53\\


\cellcolor{cyan!10}PromptSRC (DG + OSR) \cite{promptsrc} & \centering$\checkmark$ &ICCV'23 &35.72& 27.09 &24.98 &20.04& 22.02 &14.85& 30.16 &31.18&25.20
&20.44& 27.62 &  22.72  \\

\cellcolor{grey}2LM (FSDG + OSR) \cite{2lm} & \centering$\times$ &CVPR'23 & 35.22 & 21.42 &31.61&28.76& 21.30 & 13.60 & 29.73 & 34.80 & 24.50 & 17.75 & 28.47 & 23.27 \\

\cellcolor{grey}ODG-Net (ODG) \cite{odg-net} & \centering$\times$ &TMLR'23 & 34.82 & 21.67 &32.33&29.17&20.47 & 11.45 &29.16&29.40&22.05&19.08&27.77&22.15\\

\cellcolor{grey}MEDIC (ODG) \cite{medic} & \centering$\times$ &ICCV'23 & 33.91 & 21.40 &32.94&26.28& 21.31 & 11.75 & 30.35 & 33.11 & 23.73 & 19.05 &28.45&22.32\\

\cellcolor{cyan!10}\textsc{SCI-PD} (ODG)  \cite{scipd}& \centering $\checkmark$ &CVPR'24&  23.40& 25.84 &19.88&19.60& 35.27 & 44.31 & 16.95 & 19.18 & 16.25 & 23.33 &22.35&26.45\\

\cellcolor{cyan!10}\textsc{ODG-CLIP} (ODG)  \cite{odgclip}& \centering $\checkmark$ &  CVPR'24&68.89 & \cellcolor{red!20}75.56 &\cellcolor{red!20}52.43&\cellcolor{red!20}54.70& 48.69 & \cellcolor{red!20}52.93 & \cellcolor{red!20}63.74 & \cellcolor{red!20}69.53 & \cellcolor{red!20}61.05 & \cellcolor{red!20}65.50 & \cellcolor{red!20}58.96 & \cellcolor{red!20}63.64\\

\hline
\cellcolor{blue!20}\textbf{\textsc{OSLoPrompt} (Ours)} & \centering \checkmark & - &\textbf{92.71} & \textbf{94.86} &\textbf{78.89}&\textbf{76.89}& \textbf{69.73} &\textbf{64.04}& \textbf{76.30} & \textbf{74.49} & \textbf{69.00} &\textbf{67.57}&\textbf{77.32}&\textbf{75.57}\\

\bottomrule
\end{tabular}}
\label{tab_open_1}
\end{center}
\vspace*{-7.8mm}
\vspace{15pt}
\end{table*}
\begin{table*}[htbp]
\vspace*{-5mm}
\begin{center}
\scalebox{0.68}{
\begin{tabular}{lcccccccccccc|cc}
\toprule
\multicolumn{1}{l}{\multirow{2}{*}{\textbf{Methods}}} & \multicolumn{1}{c}{\multirow{2}{*}{\textbf{CLIP-based}}} & \multicolumn{1}{c}{\multirow{2}{*}{\textbf{Venue}}} 
& \multicolumn{2}{c}{\textbf{PACS}}& \multicolumn{2}{c}{\textbf{VLCS}} & \multicolumn{2}{c}{\textbf{OfficeHome}} & \multicolumn{2}{c}{\textbf{Multi-Dataset}} & \multicolumn{2}{c}{\textbf{Mini-DomainNet}} & \multicolumn{2}{|c}{\textbf{Average}}  \\
\cmidrule(lr){4-5}\cmidrule(lr){6-7}\cmidrule(lr){8-9}\cmidrule(lr){10-11}\cmidrule(lr){12-13}\cmidrule(lr){14-15}
& && \textbf{Acc} & \textbf{H-score}& \textbf{Acc} & \textbf{H-score} & \textbf{Acc} & \textbf{H-score} & \textbf{Acc} & \textbf{H-score} & \textbf{Acc} & \textbf{H-score} & \textbf{Acc} & \textbf{H-score} \\
\midrule

\cellcolor{cyan!10}CLIP + OpenMax (OSR) \cite{osr1} & \centering$\checkmark$ &CVPR'16& 68.75 & 80.98 & \cellcolor{red!20}66.25 & \cellcolor{red!20}74.74& 35.59 & 49.28 & 56.59 & 68.84 & 32.46 & 48.20& 51.93 & 64.41 \\

\cellcolor{cyan!10}CLIPN (OSR) \cite{clipn} & \centering $\checkmark$ &ICCV'23&  78.04 & 71.14 &32.92&27.95& 47.94 & 40.33 & 46.50 & 39.23 & 55.78 & 48.53 & 52.24 & 45.44
\\
\cellcolor{grey}MORGAN (FS-OSR) \cite{morgan} & \centering $\times$ &WACV'23&  46.27 & 24.06 &42.16&38.70& 36.20 & 18.63 & 35.47 & 42.80 & 37.81 & 27.06 & 39.58 &30.25\\

\cellcolor{cyan!10}StyLIP (DG + OSR)\cite{stylip} & \centering $\checkmark$ &WACV'24 & 80.10 & 70.01 &45.78	&48.93 & \cellcolor{red!20}61.87 & 42.46 & 54.58 & 49.76 & 64.03 & 60.68 & 61.27 & 54.37 \\

\cellcolor{cyan!10}PromptSRC (DG + OSR) \cite{promptsrc} & \centering$\checkmark$ &ICCV'23 &46.86 & 30.23 &36.16&32.36& 31.10 & 20.35 & 35.68 & 38.12 & 36.37 & 31.32 & 37.24 & 30.28 \\

\cellcolor{grey}2LM (FSDG + OSR) \cite{2lm} & \centering$\times$ &CVPR'23 &46.70 & 24.06 &41.67&37.36& 29.38 & 18.95 & 35.04 & 35.38 & 38.43 & 28.70 & 38.24 & 28.89 \\

\cellcolor{grey}ODG-Net (ODG) \cite{odg-net} & \centering$\times$ & TMLR'23& 46.66 & 25.92 &43.05&37.71& 34.52 & 15.96 & 34.20 & 36.93 & 39.95 & 23.72 & 39.68 & 28.05 \\

\cellcolor{grey}MEDIC (ODG) \cite{medic} & \centering$\times$ & ICCV'23 & 44.88 & 25.05 &40.53&35.56& 30.40 & 18.45 & 35.42 & 36.26 & 36.95 & 30.60 & 37.64 & 29.18 \\

\cellcolor{cyan!10}SCI-PD (ODG) \cite{scipd} & \centering $\checkmark$ &CVPR'24& 35.16 & 34.53 &30.11&30.48& 32.98 & 42.50 & 32.20 & 28.89 & 21.25 & 30.57 & 30.34 & 33.39 \\

\cellcolor{cyan!10}ODG-CLIP (ODG) \cite{odgclip} & \centering $\checkmark$ & CVPR'24 &\cellcolor{red!20}83.65 & \cellcolor{red!20}88.16 &62.93&56.89& 55.32 & \cellcolor{red!20} 49.31 & \cellcolor{red!20}74.40 & \cellcolor{red!20}76.14 & \cellcolor{red!20}74.38 & \cellcolor{red!20}65.49 & \cellcolor{red!20}70.14 &\cellcolor{red!20}67.20 \\

\hline
\cellcolor{blue!20}\textbf{\textsc{OSLoPrompt} (Ours)} & \centering $\checkmark$ & - &\textbf{93.72}&\textbf{95.01}&\textbf{79.04}&\textbf{ 77.34} &   \textbf{75.33}& \textbf{62.08 }&\textbf{ 79.75} &\textbf{80.05 }& \textbf{74.52} &\textbf{66.58 }& \textbf{80.47}  & \textbf{76.21} \\

\bottomrule
\end{tabular}}
\label{tab_open_2}
\end{center}
\vspace*{-7.8mm}
\end{table*}

\noindent \textbf{(iii) Domain-generic prompt learning:}
Our objective is to learn domain-agnostic prompts, \eg \( \mathbf{{Prompt}_{\text{gen}}^y} \) for class \( y \), covering both known classes in \( \mathcal{C} \) and the \texttt{Unknown} class in \( \mathcal{D}^{\text{aug}} \). Full context learning for \( \mathbf{{Prompt}_{\text{gen}}} \) as in \cite{coop} is insufficient, as it neglects visual cues. Additionally, since \( \mathbf{{Prompt}_{\text{gen}}} \) addresses outliers without pre-defined class names or attributes, the attribute-based conditioning of Eq. \ref{eq:ca} is ineffective. To address this, we propose a novel enrichment strategy for \( \mathbf{{Prompt}_{\text{gen}}} \).

Precisely, a subset of context tokens within \( \mathbf{{Prompt}_{\text{gen}}} \) is initialized by projecting knowledge from the learnable visual prompt, \( \mathbf{{Prompt}_v} \). Unlike existing methods \cite{cocoop} that rely on frozen per-image features extracted from $\mathcal{F}_v$ for conditioning the learnable prompts, risking the introduction of image-specific artifacts that can degrade prompt quality and affect performance (Table \ref{tab:combined_ablation}), our \( \mathbf{{Prompt}_v} \) is shared across all visual entities. This shared structure provides a more comprehensive and domain-independent source of knowledge. Formally, \( \mathbf{{Prompt}_{\text{gen}}^y} \) is defined as:
\begin{equation}
    \mathbf{{Prompt}_{\text{gen}}^y} = [\nu_{1:q}] [\text{Proj}_{vt}(\mathbf{{Prompt}_v})]_{q+1:\mathcal{M}} [\text{CLS}_y]
\end{equation}
Where \( \{\nu_{1:q}\} \) are the \( q \) directly learnable context tokens in \( \mathbf{{Prompt}_{\text{gen}}} \), and \( [\text{Proj}_{vt}(\mathbf{{Prompt}_v})]_{q+1:\mathcal{M}} \) are the tokens derived from the learnable visual prompts through the projector function \( \text{Proj}_{vt} \), and \( \mathcal{M} \) denotes the total context length, which is similar to the that of $\mathbf{{Prompt}_s}$.

Two main objectives drive the training of these domain-agnostic prompts:

\textbf{- Proposed context alignment loss:} We aim to align the context tokens of \( \mathbf{{Prompt}_{\text{gen}}} \) with those of all domain-specific prompts \( \{\overline{\mathbf{{Prompt}_s}}\}_{s=1}^{\mathcal{N}} \) over the source domains. This alignment allows \( \mathbf{{Prompt}_{\text{gen}}} \) to inherit domain-specific attributes and fine-grained semantic knowledge in the context tokens better, enabling it to generalize effectively. The alignment loss is defined as:
\begin{multline}
    \mathcal{L}_{\text{align}} = \underset{\substack{\{\nu_{1:q}\}, \mathbf{w^q}, \mathbf{w^k}, \mathbf{w^v}, \\
    \text{Proj}_{vt}, \mathbf{{Prompt}_v}}}{\min} \; \;\underset{s=1}{\overset{\mathcal{N}}{\sum}} \; \underset{(x^s, y^s) \in \mathcal{P}(\mathcal{D}_s)}{\mathbb{E}} \;  \; \Big[ 1 - \\ \operatorname{cosine}  \big( \mathcal{F}_t
    (\mathbf{{Prompt}_{\text{gen}}^{y^s}}), 
    \mathcal{F}_t
    (\overline{\mathbf{{Prompt}_s^{y^s}}}(x^s)) \big) \Big]
\end{multline}

\textbf{- Supervised visual-textual contrastive loss:} We further refine the domain-agnostic prompts using a supervised visual-textual contrastive framework, supplemented by a cross-entropy loss \( \mathcal{L}_{\text{ce}}^{\text{dom-gen}} \) over the augmented dataset \( \mathcal{D}^{\text{aug}} \) similar to Eq. \ref{eq:ce}. This training aims to enhance the separation between known classes and pseudo-open samples. 

The total loss to train \textsc{OSLoPrompt} is,

\begin{equation}\centering
    \mathcal{L}_{\text{total}} = \mathcal{L}_{\text{ce}}^{\text{dom-gen}} + \mathcal{L}_{\text{ce}}^{\text{dom-spec}} + \mathcal{L}_{\text{align}}
\end{equation}

During inference, for an input \( x^t \) from the target domain \( \mathcal{D}_t \), the model classifies \( x^t \) by maximizing the cosine similarity between \( \mathcal{F}_v(x^t) \) and \( \mathbf{{Prompt}^{y^t}_{\text{gen}}} \):
\begin{equation}
    \centering
    \overline{y}^{t} = \underset{y^t \in \mathcal{C} \cup \texttt{Unknown}}{\text{argmax}} p(y^t|x^t, \mathcal{F}_v, \mathcal{F}_t, \mathbf{{Prompt}_{\text{gen}}^{y^t}})
\end{equation}

\textbf{Pseudo-code} of the pipeline is mentioned in \textbf{Sup Mat}.
 \section{Experimental Evaluations}
\noindent{\textbf{Datasets:}} We evaluate \textsc{OSLoPrompt} on five benchmark datasets: Office-Home \cite{officehome}, PACS \cite{pacs1}, VLCS \cite{vlcs}, Mini-DomainNet \cite{domainnet}, and Multi-Dataset \cite{daml}, following standard known-novel class splits \cite{daml, odgclip}, but considering one and five training samples per class. Further details regarding the splits are mentioned in the \textbf{Sup Mat}. We also consider the ImageNet suite \cite{coop, cocoop} for evaluating the closed-set DG performance in the low-shot setting.

\noindent{\textbf{Architecture details:}} For all CLIP-based models including ours, ViT-B/32 is employed as the visual backbone $\mathcal{F}_v$, with a Transformer \cite{transformer} serving as $\mathcal{F}_t$. We use the official implementations for the available methods, while implementing \cite{2lm} on our own. For \cite{morgan}, which was originally designed with 3D-CNN, we re-implement with the ResNet-50 based backbone \cite{resnet} to accommodate the RGB datasets.

\noindent{\textbf{Training and evaluation:}} Training is performed over 10 epochs with the AdamW optimizer \cite{loshchilov2017decoupled}. Batch sizes are set per dataset: 6 for PACS/VLCS, 9 for Office-Home, Multi-Dataset, and Mini-DomainNet, with each batch incorporating three pseudo-open samples from each source domain. All CLIP-based methods use a textual prompt context length  $\mathcal{M}=4$ and a visual prompt context length $m=2$. In $\mathbf{{Prompt}_{\text{gen}}}$, two tokens are initialized from $\mathbf{{Prompt}_v}$, while the other two are initialized randomly. $\text{Proj}_{vt}$ is implemented as a meta-net similar to the one in \cite{cocoop}, with $\mathcal{B}=4$ attributes per class added into $\mathcal{A}$.
For domain-specific prompts, domain names are used directly where available. For VLCS and Multi-Dataset, prompts are formatted as \texttt{[Photo] of a [CLS]}. We filter out images with very low entropy of the grey-value distributions ($\leq 0.2)$ as these images are irrelevant.
We evaluate using two metrics under the leave-one-domain-out protocol \cite{odgclip}: \textbf{a)} top-1 accuracy (Acc) for closed-set classes, and \textbf{b)} the harmonic mean (H-score) for combined closed-set and open-set performance. Results are averaged over three runs.

\begin{table}[htbp]
\caption{Comparative analysis of \textbf{1-shot closed-set DG performance on the ImageNet benchmark} using ViT-B/32.}\vspace{-5mm}
\begin{center}
\scalebox{0.65}{
\begin{tabular}{l|c|cccc|c}
\toprule
\textbf{Methods} & \textbf{Source} & \multicolumn{4}{c|}{\textbf{Target}} & \textbf{Average} \\
\cmidrule(lr){2-2} \cmidrule(lr){3-6} \cmidrule(lr){7-7}
 & \textbf{ImageNet} & \textbf{IN-V2} & \textbf{IN-Sketch} & \textbf{IN-A} & \textbf{IN-R} & \textbf{Average} \\
\midrule
CoOp \cite{coop} & 63.8 & 56.5 & 41.4 & 31.4 & 66.8 & 49.02 \\
CoCoOp \cite{cocoop} & \cellcolor{red!20}\textbf{64.4} & 56.4 & 41.0 & \cellcolor{red!20}\textbf{32.1} & 66.7 & 49.05 \\
KgCoOp \cite{kgcoop} & 64.2 & 56.2 & \cellcolor{red!20}41.7 & 31.4 &\cellcolor{red!20} 67.7 & 49.25 \\
MaPLe \cite{maple} & 64.2 &\cellcolor{red!20}\textbf{56.7 }& \cellcolor{red!20}41.7 & 32.0 & \cellcolor{red!20}67.7 & \cellcolor{red!20}49.52 \\
StyLIP \cite{stylip} & 64.0 & 56.6 & 40.9 & 31.7 & 66.7 & 48.98 \\
ODG-CLIP \cite{odgclip} & 61.3 & 53.1 & 38.7 & 28.9 & 63.5 & 46.05 \\
\rowcolor{blue!20}\textbf{\textsc{OSLoPrompt}} & 64.2 & 56.5 & \textbf{42.0} & \textbf{32.1} & \textbf{68.2} & \textbf{49.70} \\
\bottomrule
\end{tabular}
}
\end{center}
\label{tab:imagenet}
\end{table}

\noindent{\textbf{Competing methods:}} Given the absence of existing LSOSDG methods in the literature, we design the following baselines:
 \textbf{Open-Set Recognition (OSR) and Few-Shot OSR methods:} We include CLIP + OpenMax \cite{osr1}, CLIPN \cite{clipn}, and MORGAN \cite{morgan}.
\textbf{FSDG methods:} We integrate the FSDG method \cite{2lm} with OSR \cite{osr1} capabilities.
\textbf{CLIP-based closed-set DG + OSR methods:} This includes PromptSrc \cite{promptsrc} and \textsc{StyLIP} \cite{stylip} with an \texttt{Unknown}-class prompt like ours\footnote{The conventional prompting techniques like \cite{coop, cocoop, maple, kgcoop} overfit the known classes, failing in general on the open-set detection. We do not report them in the tables.}.
\textbf{Existing ODG techniques:} We evaluate both CLIP and non-CLIP based methods such as \cite{odg-net, medic, odgclip, scipd}\footnote{We did not compare with \cite{peng2024advancing} as it requires training samples from all classes in all domains, unlike our setting that has no such constraint.}.
In the \texttt{ImageNet} experiments, we benchmark against existing CLIP-based prompting techniques \cite{coop, cocoop, maple, kgcoop, stylip, odgclip} More details on the implementations are mentioned in \textbf{Sup Mat}.

\begin{table}[htbp]
\centering
\caption{\textbf{Ablation analysis} for Office-Home (O.H.) and Mini-DomainNet (M.DNet) in a 1-shot setting on H-score. Details about these implementations are mentioned in \textbf{Sup Mat}.}
\resizebox{0.45\textwidth}{!}{\begin{tabular}{>{\raggedright\arraybackslash}p{6.5cm}cc}
\toprule
\textbf{Methods} & \textbf{O.H.} & \textbf{M.DNet} \\
\midrule
\rowcolor{blue!10} \multicolumn{3}{c}{\textbf{Analysis of domain-specific prompts}} \\
$\checkmark$ Manual prompting: \texttt{Domain of a CLS} & 59.33 & 65.88 \\
$\checkmark$ Manual prompting with image conditioning & 62.96 & 67.27 \\
$\checkmark$ Manual prompting expanded with ad-hoc attributes from $\mathcal{A}$  \cite{attr1}& 60.69 & 63.82 \\
$\checkmark$ Manual prompting with ad-hoc attributes and image conditioning & 62.16 & 65.11 \\
$\checkmark$ Visual attributes learning \cite{kim2024aapl} & 58.35 & 60.57 \\
$\checkmark$ \textbf{Proposed cross-attention approach} & \textbf{64.04} & \textbf{67.57} \\
\addlinespace
\rowcolor{blue!10} \multicolumn{3}{c}{\textbf{Analysis of domain-agnostic prompts}} \\
$\checkmark$ Full context learning \cite{coop} & 60.81 & 53.43 \\
$\checkmark$ Image-cond. context learning \cite{cocoop} & 63.10 & 59.61 \\
$\checkmark$ \textbf{Proposed multi-modal prompting} & \textbf{64.04} & \textbf{67.57} \\
\addlinespace
\rowcolor{blue!10} \multicolumn{3}{c}{\textbf{Sensitivity to the number of attributes per class in $\mathcal{A}$}} \\
$\checkmark$ \textbf{4} & \textbf{64.04} & \textbf{67.57} \\
$\checkmark$ 8 & 63.97 & 66.36 \\
$\checkmark$ 12 & 63.79 & 65.14 \\
\addlinespace
\rowcolor{blue!10} \multicolumn{3}{c}{\textbf{Importance of the loss terms}} \\
$\checkmark$ $\mathcal{L}_{\text{ce}}^{\text{dom-gen}}$ (no domain-specific guidance) & 62.51 & 63.86 \\
$\checkmark$ $\mathcal{L}_{\text{ce}}^{\text{dom-gen}} + \mathcal{L}_{\text{ce}}^{\text{dom-spec}}$ (partial domain-specific guidance) & 62.52 & 65.56 \\
$\checkmark$ $\boldmath{\mathcal{L}_{\text{ce}}^{\text{dom-gen}} + \mathcal{L}_{\text{ce}}^{\text{dom-spec}} + \mathcal{L}_{\text{align}}}$ & \textbf{64.04} & \textbf{67.57} \\
\addlinespace
\rowcolor{blue!10} \multicolumn{3}{c}{\textbf{Pseudo-open image synthesis}} \\
$\checkmark$ Generic sample generation of \cite{odgclip}& 41.09 & 49.07 \\
$\checkmark$ Mixup-based \cite{cumix} pseudo-open images & 57.26 & 64.85 \\
$\checkmark$ \textbf{Our fine-grained sample generation}& \textbf{64.04} & \textbf{67.57} \\
\bottomrule
\end{tabular}}
\label{tab:combined_ablation}
\end{table}

\subsection{Discussions on the main results}
Tables \ref{tab_open_1} shows that \textsc{OSLoPrompt} consistently outperforms all competing methods in 1-shot and 5-shot settings. Non-CLIP-based models \cite{odg-net, 2lm, medic, morgan} struggle with closed-set accuracy and open-sample classification, leading to lower H-scores. While CLIP-based methods \cite{clipn, stylip, promptsrc, scipd} improve on these metrics, they still face challenges in generalization and open-set detection, with \textsc{StyLIP} achieving the highest H-scores among them at 44.53\% (1-shot) and 54.37\% (5-shot). ODG-CLIP \cite{odgclip}, the strongest competitor, achieves H-scores of 63.64\% (1-shot) and 67.20\% (5-shot), yet still lags behind \textsc{OSLoPrompt}.

\textsc{OSLoPrompt} attains average H-scores of 75.57\% (1-shot) and 76.21\% (5-shot) across five datasets, outperforming ODG-CLIP by $11.93 \%$ (1-shot) and $9.01 \%$ (5-shot), respectively. Although 5-shot performance generally improves, in Mini-DomainNet and Office-Home, open-set performance dips slightly, likely due to fine-grained distinctions causing mild overfitting, causing the H-score to decrease by $1-2 \%$ than those of the 1-shot cases. Nonetheless, these consistent gains underscore \textsc{OSLoPrompt}'s superior generalization with limited data.

Additionally, we evaluated various prompting methods on the ImageNet suite \cite{coop, cocoop} for single-source multi-target DG, using a single training sample per class from \texttt{ImageNet} and testing on \texttt{ImageNet-v2} \cite{imagenetv2}, \texttt{ImageNet-Sketch} \cite{imagenetsketch}, \texttt{ImageNet-A} \cite{imageneta}, and \texttt{ImageNet-R} \cite{imagenetr}. This closed-set evaluation used a generic domain-specific prompt \texttt{[Photo] of a [CLS]}, limiting the advantages of our context distillation objective $\mathcal{L}_{\text{align}}$. Notwithstanding this fact, as shown in Table \ref{tab:imagenet}, \textsc{OSLoPrompt} outperforms competing methods in three of four target domains, with an average performance of 49.70\%, 0.18\% higher than the second-best method.
\begin{figure*}
    \centering
    \includegraphics[width=\linewidth]{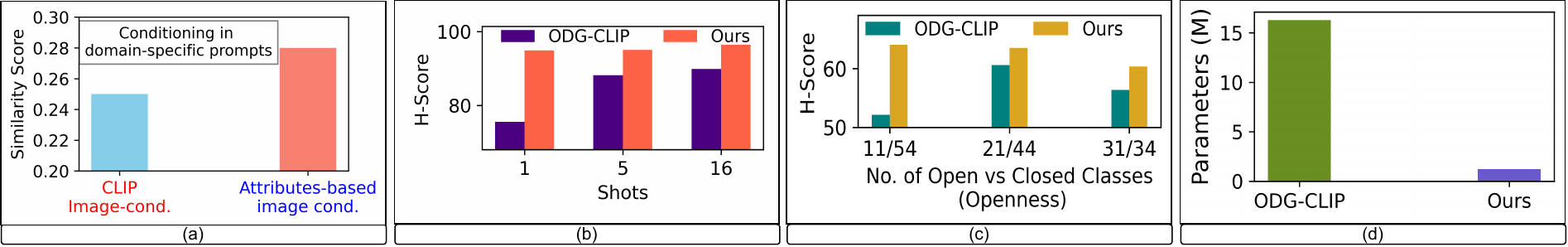}\vspace{-2mm}
    \caption{\textbf{(a)} Comparison of the \textbf{cosine similarity} between image and prompt embeddings from $\mathcal{F}_v$ and $\mathcal{F}_t$ under CLIP image feature conditioning and our proposed semantic attribute-driven encoding on the domain-specific prompts on PACS, showing improved image-prompt alignment with our approach. \textbf{(b)} \textbf{Sensitivity} of ODG-CLIP \cite{odgclip} and \textsc{OSLoPrompt} \textbf{to the number of training samples per class} on PACS. \textbf{(c)} \textbf{Openness sensitivity} of \textsc{OSLoPrompt} and ODG-CLIP in the 1-shot Office-Home case for different known and novel class ratios. \textbf{(d)} Comparison of \textbf{trainable} parameters between ODG-CLIP and our method.}
    \label{fig:ablation_new}
\end{figure*}
\begin{figure*}
    \centering
    \vspace*{-2mm}
    \includegraphics[width=\linewidth]{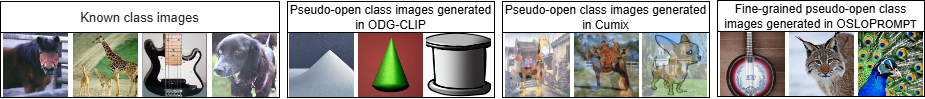}
    \vspace*{-5 mm}
    \caption{\textbf{Pseudo-open images} generated by ODG-CLIP \cite{odgclip} are highly \textbf{coarse-grained} in relation to the known classes. While CuMix \cite{cumix} provides improved fine-grained details compared to ODG-CLIP, it still lacks proper semantic coherence. Our pseudo-open image generation achieves a \textbf{fine-grained} level of detail, maintaining both semantic relevance and class-specific granularity (for PACS).}
    \label{fig:images_fine}
    \vspace*{-4.5mm}
\end{figure*}

\subsection{Main ablation analysis}
Our comprehensive ablation analysis, outlined in Table \ref{tab:combined_ablation}, evaluates the impact of various components of our methodology across two datasets, Office-Home and Mini-DomainNet, in a 1-shot scenario. Further ablation analysis, specifically \textbf{analysis of the context lengths} and visualizations, are mentioned in the \textbf{Sup Mat}.

\textbf{- Analysis of domain-specific prompts:} To validate our attribute-enriched domain-specific prompts, we compared them to several alternatives: manual static prompts, expanded static prompts with class-wise attributes from $\mathcal{A}$ following \cite{attr1}, static prompts augmented with image features from $\mathcal{F}_v$, image conditioning within attribute-based static prompts, and the integration of visual attribute learning \cite{kim2024aapl}. Our approach, which aligns visual embeddings with weighted semantic attributes, outperformed all others, exceeding the next best (image-conditioned static prompts) by $1.08\%$ on Office-Home and $0.30\%$ on Mini-DomainNet. Fig.~\ref{fig:ablation_new} (a) illustrates the qualitative improvements of attribute guidance over full image-feature conditioning for image-prompt alignment.

\textbf{- Analysis of domain-agnostic prompts:} Among domain-agnostic prompt learning methods, our strategy—combining contexts from learnable visual prompts with directly learnable tokens—outperforms both full-context learning from initialization ``\texttt{Photo of a}'' \cite{coop} and image-conditioned full-context learning of \cite{cocoop}. This highlights the benefit of embedding generic visual insights into agnostic prompts, yielding approximately $8\%$ improvement on Mini-DomainNet on H-score.

\textbf{- Attribute count per class in $\mathcal{A}$:} Optimizing the attribute count to four per class in $\mathcal{A}$ balances detail and model complexity, outperforming configurations with eight and twelve attributes. For Mini-DomainNet, four attributes yield $1.21\%$ better performance than eight.

\textbf{- Sensitivity to loss terms:} Excluding $\mathcal{L}_{\text{ce}}^{\text{dom-spec}}$ and $\mathcal{L}_{\text{align}}$ from the total loss results in nearly an $3.71\%$ H-score drop on Mini-DomainNet, underscoring domain-specific guidance's importance. Including $\mathcal{L}_{{ce}}^{\text{dom-spec}}$ alongside $\mathcal{L}_{\text{ce}}^{\text{dom-gen}}$, thus enabling domain-specific gradient updates for $\mathbf{{Prompt}_v}$, enhances performance by about $1.7\%$ over the baseline. $\mathcal{L}_{\text{align}}$ further improves results by $2.01\%$.

   
\textbf{- Sensitivity to varying shots and known-to-novel class ratios:} As shown in Fig. \ref{fig:ablation_new} (b), our method sustains high H-scores as the number of training samples per class increases from 1 to 16, beating ODG-CLIP consistently. Additionally, Fig. \ref{fig:ablation_new} (c) illustrates the sensitivity of ODG-CLIP and \textsc{OSLoPrompt} to varying ratios of open-to-closed classes, or \textit{openness}. Our method consistently surpasses ODG-CLIP across a broad range of openness.

\textbf{- Pseudo-open image generation (Fig.~\ref{fig:images_fine}):} Replacing our pseudo-open image generation with the coarse-level approach from \cite{odgclip} led to a 22.95\% and 18.5\% decrease in H-score across both datasets, highlighting the importance of a precise closed-open decision boundary in LSOSDG. While CuMix \cite{cumix}, which combines random image regions to form pseudo-open samples, produces finer results than \cite{odgclip} given the fact that the mixed images are not wildly divergent from the originally known set of classes, it still trails our approach by 6.78\% and 2.72\% in H-score. More visual results are shown in \textbf{Sup Mat}.

\textbf{- Model complexity:} Fig.~\ref{fig:ablation_new} (d) illustrates the computational efficiency of \textsc{OSLoPrompt} compared to ODG-CLIP. Overall, ODG-CLIP has 142M model parameters, contrary to 127M of ours. Amongst them, 16.3M parameters are trainable for ODG-CLIP while it is 1.21M for ours.

\vspace{-2mm}
\section{Conclusions}\vspace{-2mm}
This paper introduces the LSOSDG problem setting and presents \textsc{OSLoPrompt} as an effective solution. Our approach integrates a domain-agnostic prompt learning strategy with structured regularization through ad-hoc domain-specific prompts, enriched by image-to-semantic attribute encoding and generic visual prompts, enabling low-shot generalization and robust outlier detection. To enhance fine-grained outlier detection, we introduce a controlled mechanism for synthesizing pseudo-open samples by selectively querying diffusion models. \textsc{OSLoPrompt} consistently outperforms other methods across benchmarks. Future work will extend \textsc{OSLoPrompt} to broader open-world applications, including structured prediction tasks.
\vspace{-1mm}
\section{Acknowledgements}\vspace{-2mm}
The authors sincerely acknowledge the support from AWL Japan and Adobe Research India.







\section{Contents of the supplementary}

In the supplementary materials, we list the following information:

\begin{enumerate}
    \item Detailed descriptions of the datasets, including known and novel class splits in Section \ref{sec:dataset}.
     \item List of variable names used in the paper and their purpose is detailed in Section \ref{sec:variables}.
    \item Pseudo-code representation of the proposed method in Section \ref{pseudo}.
    \item Ablation study examining the impact of context length in Section \ref{context}.
    \item Comprehensive literature survey on \textit{classification with attributes in CLIP} in Section \ref{sec:survey}.
        \item List of attributes and generated pseudo-open class names in Table \ref{tab:attribute} and \ref{tab:fine_grained_samples} respectively.
    \item Implementation details for both comparative and ablation methods in Section \ref{sec:implementation}.
    \item Additional visualizations of the generated pseudo-open images in Figure \ref{fig:grid} (Section \ref{sec:vis}).
    \item Expanded tables detailing domain combinations across the five datasets in Table   \ref{tab:minidomainnet_domain_performance}, \ref{tab:multidataset_domain_performance}, \ref{tab:officehome_domain_performance}, \ref{table:vlcs_domain_performance} and \ref{table:pacs_domain_performance}.
    \item Few limitations of \textsc{OSLoPrompt} in Section \ref{sec:limitations}.
\end{enumerate}

\begin{table*}[h]
\centering
\caption{Summary of the datasets used.}
\label{tab:datasets}
\resizebox{0.8\textwidth}{!}{%
\begin{tabular}{@{}>{\bfseries}lccc@{}}
\toprule
\rowcolor[HTML]{EFEFEF} \textbf{Dataset} & \textbf{Images} & \textbf{Classes} & \textbf{Domains} \\ \midrule
Office-Home \cite{officehome}   & 15,500   & 65         & 4 (Art, Clipart, Product, Real) \\
PACS \cite{pacs1}                & 9,991    & 7          & 4 (Artpaint, Cartoon, Sketch, Photo) \\
Multi-Dataset\cite{multidataset} & Combined & 20 open    & Various (Office-31 \cite{office31}, STL-10\cite{stl}, VisDA2017\cite{visda}, DomainNet\cite{domainnet}) \\
Mini-DomainNet \cite{domainnet}  & 362,470  & 125        & 4 (Clipart, Painting, Real, Sketch) \\
VLCS \cite{vlcs}                & 10,729   & 5          & 4 (PASCAL VOC 2007\cite{pascal}, Caltech\cite{caltech}, LabelMe\cite{labelme}, Sun\cite{sundata}) \\ \bottomrule
\end{tabular}%
}
\end{table*}

\begin{table*}[htbp!]
\centering
\caption{Known-novel class splits for the ODG settings: PACS, VLCS, Office-Home, Multi-dataset, and Mini-DomainNet datasets. Indices follow the alphabetical order of the class names.}
\label{tab:split}
\vspace*{-3mm}
\begin{adjustbox}{width=\linewidth}
\begin{tabular}{@{}>{\bfseries}lcccccc@{}}
\toprule
\rowcolor[HTML]{EFEFEF} \textbf{Domain} & \textbf{PACS} & \textbf{VLCS} & \textbf{Office-Home} & \textbf{Multi-Datasets } & \textbf{Mini-DomainNet } \\ 
\midrule
Source 1 & 3, 0, 1       & 0, 1       & 0–14, 21–31         & 0–30           & 0–19, 40–59       \\ 
Source 2 & 4, 0, 2       & 1, 2       & 0–8, 15–20, 32–42   & 1, 31–41       & 0–9, 20–39, 80–89 \\ 
Source 3 & 5, 1, 2       & 2, 3       & 0–2, 9–20, 43–53    & 31, 33–34, 41–47 & 10–19, 40–49, 60–79 \\ 
\midrule
Target   & 0–6           & 0–4        & 0, 3–4, 9–10, 15–16, 21–23, 32–34, 43–45, 54–64 & 0, 1, 5–6, 10–11, 14, 17, 20, 26, 31–36, 39–43, 45–46, 48–67 & 0–4, 8–17, 25–34, 43–47, 75–79, 83–87, 90–125 \\ 
\bottomrule
\end{tabular}
\end{adjustbox}
\vspace{-2mm}
\end{table*}

\section{Datasets descriptions}
\label{sec:dataset}
\textbf{Office-Home Dataset} \cite{officehome}:  
The Office-Home dataset comprises \textbf{15,500 images}, carefully organized into \textbf{65 distinct classes} that span across four visually diverse domains: \textit{Art}, \textit{Clipart}, \textit{Product}, and \textit{Real}. Each domain represents a unique visual style, ranging from artistic renderings to photographic images, making the dataset highly valuable for evaluating domain adaptation and transfer learning models. This dataset is particularly suited for domain generalization, multi-domain learning, and visual recognition tasks.

\textbf{PACS Dataset} \cite{pacs}:  
The PACS (Photo, Artpaint, Cartoon, Sketch) dataset consists of \textbf{9,991 images}, categorized into \textbf{seven broad classes}: \textit{Dog}, \textit{Elephant}, \textit{Giraffe}, \textit{Guitar}, \textit{House}, \textit{Horse}, and \textit{Person}. These images are drawn from four distinct domains: \textit{Artpaint}, \textit{Cartoon}, \textit{Sketch}, and \textit{Photo}, representing varying styles and abstraction levels. The dataset is widely recognized for its benchmark utility in domain generalization research, especially for testing models' robustness to domain shifts.

\textbf{VLCS Dataset}\cite{vlcs}: This dataset combines images from four different datasets namely (PASCAL VOC 2007 \cite{pascal}, Caltech \cite{caltech}, LabelMe \cite{labelme} and Sun  \cite{sundata}) consisting of images spread across five categories namely Bird, Car, Chair, Dog, and Person. We consider four categories as closed-set and the remaining category as open-set. Each of the datasets is considered as a separate domain.

\textbf{Multi-Dataset} \cite{multidataset}:  
The Multi-Dataset combines data from several prominent public datasets, including \textit{Office-31} \cite{office31}, \textit{STL-10} \cite{stl}, and \textit{VisDA2017} \cite{visda}. Additionally, it incorporates four domains from \textit{DomainNet} \cite{domainnet}, resulting in a richly diverse dataset. This composite dataset includes \textbf{20 open classes}, intentionally absent from the joint label set of the source domains. This design facilitates tasks such as \textit{open-set domain adaptation}, where models are challenged to handle unseen categories and cross-domain learning, providing a comprehensive benchmark for domain adaptation techniques.

\textbf{Mini-DomainNet} \cite{domainnet}:  
The Mini-DomainNet is a compact yet diverse subset of the DomainNet dataset, featuring images from \textbf{125 categories} across \textbf{four domains}: \textit{Clipart}, \textit{Painting}, \textit{Real}, and \textit{Sketch}. Each domain reflects a distinct visual characteristic, offering a balanced data distribution to evaluate models in multi-domain and transfer learning scenarios. This smaller-scale dataset is optimized for quick experimentation while maintaining the challenge and diversity of the full DomainNet.

Table \ref{tab:datasets} shows the dataset details, while Table \ref{tab:split} details the known-novel class splits, following \cite{daml, odg-net, odgclip}. In Table \ref{tab:split}, the class names are indexed to integers alphabetically.

\section{Variable names and their purpose}\label{sec:variables}

Table \ref{tab:variables} details the same.
\begin{table*}[!ht]
\centering
\renewcommand{\arraystretch}{1.2}
\setlength{\tabcolsep}{5pt}
\caption{Variables used in the \textsc{OSLoPrompt} framework.}
\label{tab:variables}
\begin{tabular}{|>{\centering\arraybackslash}m{0.25\textwidth}|>{\centering\arraybackslash}m{0.7\textwidth}|}
\hline
\rowcolor[HTML]{EFEFEF} \textbf{Variable} & \textbf{Description} \\ \hline
\multicolumn{2}{|c|}{\cellcolor[HTML]{D9EAD3}\textbf{Dataset and domains}} \\ \hline
$\mathcal{D}$ & Source domains. \\
$\mathcal{D}_s$ & $s^{th}$ source domain. \\
$\mathcal{X}_s, \mathcal{Y}_s$ & Input images and labels in the $s^{th}$ source domain. \\
$\mathcal{C}$ & Combined set of classes across all the source domains. \\
$\mathcal{D}_t, \mathcal{X}_t, \mathcal{Y}_t$ & Target domain dataset, inputs, and labels. \\ \hline
\multicolumn{2}{|c|}{\cellcolor[HTML]{F4CCCC}\textbf{Target domain class definitions}} \\ \hline
$\mathcal{Y}_t^{\text{known}}$ & Known target domain classes. \\
$\mathcal{Y}_t^{\text{novel}}$ & Novel or outlier classes. \\ \hline
\multicolumn{2}{|c|}{\cellcolor[HTML]{CFE2F3}\textbf{Data augmentation}} \\ \hline
$\mathcal{C}^{\text{open}}$ & Synthesized pseudo-open class names by GPT-4o. \\
$\mathcal{D}^{\text{open}}$ & Synthesized pseudo-open images by Stable Diffusion. \\
$\mathcal{D}^{\text{aug}}$ & Augmented dataset: $\mathcal{D} \cup \mathcal{D}^{open}$. \\ \hline
\multicolumn{2}{|c|}{\cellcolor[HTML]{FFF2CC}\textbf{Prompts and attributes}} \\ \hline
$\mathbf{{Prompt}_v}$ & Learnable visual prompts at the first ViT layer of $\mathcal{F}_v$. \\
$\mathbf{{Prompt}_s^{y^s}}$ & Domain-specific static prompts. \\
$\mathcal{A}_{y^s}, \mathcal{A}'_{y^s}(x^s)$ & GPT-4o generated class-wise attributes and attribute-enhanced image embeddings through cross-attention. \\
$\mathcal{A}''(x^s)$ & Class-agnostic semantic encodings for the images. \\
$\overline{\mathbf{{Prompt}_s^{y^s}}}(x^s)$ & Final dynamic domain-specific prompt conditioned on the image. \\ 
$\mathbf{{Prompt}_{\text{gen}}^y}$ & Domain-agnostic prompts. \\ \hline
\multicolumn{2}{|c|}{\cellcolor[HTML]{D9D2E9}\textbf{Training objectives}} \\ \hline
$\mathcal{L}_{\text{ce}}^{\text{dom-spec}}$ & Supervised contrastive loss for domain-specific prompts. \\
$\mathcal{L}_{\text{align}}$ & Context alignment loss. \\
$\mathcal{L}_{\text{ce}}^{\text{dom-gen}}$ & Supervised contrastive loss for domain-agnostic prompts. \\
$\mathcal{L}_{\text{total}}$ & Total loss combining all objectives. \\ \hline
\multicolumn{2}{|c|}{\cellcolor[HTML]{EAD1DC}\textbf{Model components}} \\ \hline
$\mathcal{F}_v, \mathcal{F}_t$ & CLIP visual and textual encoders. \\
$\text{Proj}_{vt}$ & Projector to transform the visual prompts onto the subset of context tokens of the domain-agnostic prompts. \\
$\mathbf{w_k}, \mathbf{w_v}, \mathbf{w_q}$ & Projections for the query, key, and value for cross-attention: $\mathcal{F}_v^q = \mathcal{F}_v \mathbf{w_q}^T$, $\mathcal{F}_t^k = \mathcal{F}_t \mathbf{w_k}^T$, $\mathcal{F}_t^v = \mathcal{F}_t \mathbf{w_v}^T$. \\ \hline
\end{tabular}
\end{table*}

\section{Pseudo-code of our training process}\label{pseudo}
We detail the training process of \textsc{OSLoPrompt} in Algorithm \ref{alg:oslprompt}, using the variable names from the main paper.

\begin{algorithm}

\caption{\textsc{OSLoPrompt}: Training algorithm for obtaining domain-agnostic prompts capable of solving LSOSDG}
\label{alg:oslprompt}
\begin{algorithmic}[1]

\Require Training data $\mathcal{D}$, domains $\{s\}_{s=1}^\mathcal{N}$, the synthesized pseudo-open dataset $\mathcal{D}^{\text{open}}$: $\mathcal{D}^{\text{aug}} = \mathcal{D} \cup \mathcal{D}^{\text{open}}$, $\mathcal{F}_v$, $\mathcal{F}_t$
\Ensure Optimize parameters $\{\nu_{1:q}\}, \mathbf{w^q}, \mathbf{w^k}, \mathbf{w^v}, \text{Proj}_{vt}, \mathbf{{Prompt}_v}$
    \State Initialize $\mathbf{{Prompt}_v}$ for the visual prompt learning in $\mathcal{F}_v$
        \State Construct and initialize the domain-agnostic prompt $\mathbf{{Prompt}_{\text{gen}}}$ as given in Eq \textcolor{red}{9}

\While{training not converged}
    \State Sample a batch $\{(x, y)\}$ from $\mathcal{D}^{\text{aug}}$
    \For{$s = 1$ to $\mathcal{N}$}
        \State Extract samples $\{(x^s, y^s)\}$ belonging to $\mathcal{D}_s$
        \State Initialize the domain-specific static prompt $\mathbf{{Prompt}_s^{y^s}}$ of $\mathcal{D}_s$ Eq \textcolor{red}{3}
        \State Compute the attribute-enhanced embedding $\mathcal{A}'_{y^s}(x^s)$ using Eq \textcolor{red}{4} through the notion of cross attention using query-key-value formulation
        \State Class agnostic encoding 
        $\mathcal{A}''(x^s)$ is computed by averaging across all the classes Eq \textcolor{red}{5}
        \State Updated image-driven semantic attributes conditioned domain specific prompt $\overline{\mathbf{{Prompt}_s^{y^s}}}(x^s)$ is obtained from $\mathbf{{Prompt}_s^{y^s}}$ and $\mathcal{A}''(x^s)$ Eq \textcolor{red}{6}
        \State Compute the class-posterior probability  $p(y^s | x^s)$ using Eq \textcolor{red}{8} and $\mathcal{L}_{\text{ce}}^{\text{dom-spec}}$ using Eq \textcolor{red}{7}
    \EndFor
    \State $\mathcal{L}_{\text{align}}$ is obtained by computing cosine similarity of  \( \mathbf{{Prompt}_{\text{gen}}} \) and \( \{\overline{\mathbf{{Prompt}_s}}\}_{s=1}^{\mathcal{N}} \) Eq \textcolor{red}{10} for the known classes in $\mathcal{C}$
    \State $\mathcal{L}_{\text{ce}}^{\text{dom-gen}}$ is calculated given $(x,y) \in \mathcal{D}^{\text{aug}}$ for the classes $\mathcal{C} \cup \text{Unknown}$ 
    \State Training objectives: $\mathcal{L}_{\text{total}} \gets \underset{\substack{\{\nu_{1:q}\}, \mathbf{w^q}, \mathbf{w^k}, \mathbf{w^v}, \\
    \text{Proj}_{vt}, \mathbf{{Prompt}_v}}} {\min} \; \; \Bigg[\mathcal{L}_{\text{align}} + \mathcal{L}_{\text{ce}}^{\text{dom-spec}} + \mathcal{L}_{\text{ce}}^{\text{dom-gen}} \Bigg]$  (Eq. \textcolor{red}{11})
\EndWhile

\end{algorithmic}
\end{algorithm}

\section{Ablation on the context lengths}\label{context}
In this particular section, we analyze the effect of the number of directly learnable tokens $q$ that are introduced in addition to tokens derived from the visual prompts as given in Eq \textcolor{red}{9}. As observed in Table \ref{tab:q}, we can see that when $q=2$, it leads to the best harmonic score. This highlights the need for the balance of the directly learnable context tokens and tokens derived from visual prompts. In Table \ref{tab:context}, we can see that there is better H-score for the context length $\mathcal{M}=8$ for PACS, but we follow context length 4 since it is followed majorly in the literature including \cite{odgclip}, giving better results on majority of the datasets.
\begin{table*}[h]
\centering
\caption{Ablation on PACS dataset for the number ($q$) of directly learnable context tokens in the domain-agnostic prompts when the total context length $\mathcal{M}$ is 4.}
\label{tab:q}
\vspace{-3mm}
\begin{adjustbox}{width=0.25\textwidth}
\begin{tabular}{@{}cc@{}}
\toprule
\rowcolor[HTML]{EFEFEF} \textbf{Number of Tokens ($q$)} & \textbf{H-score} \\ 
\midrule
0 & 94.32 \\ 
1 & 94.74 \\ 
2 & \textbf{94.86} \\ 
3 & 93.47 \\ 
4 & 92.15 \\ 
\bottomrule
\end{tabular}
\end{adjustbox}
\vspace{-2mm}
\end{table*}

\begin{table*}[h]
\centering
\caption{Ablation on context length for the PACS dataset\cite{pacs} 1-shot setting.}
\label{tab:context}
\vspace{-3mm}
\begin{adjustbox}{width=0.25\textwidth}
\begin{tabular}{@{}cc@{}}
\toprule
\rowcolor[HTML]{EFEFEF} \textbf{Context Length ($\mathcal{M}$)} & \textbf{H-score} \\ 
\midrule
4  & 94.86 \\ 
8  & \textbf{95.44} \\ 
16 & 94.21 \\ 
\bottomrule
\end{tabular}
\end{adjustbox}
\vspace{-2mm}
\end{table*}
\section{Literature survey on prompting with descriptions in CLIP}
\label{sec:survey}
The classification accuracy of CLIP on downstream tasks and open-vocabulary datasets is highly influenced by the quality of text prompts~\cite{radford2019language}. Prior works have explored this sensitivity through simple handcrafted templates (\eg, ``a photo of a [\texttt{CLS}]'')~\cite{radford2019language} or by augmenting these templates with semantically richer attributes generated by large language models (LLMs)~\cite{attr2}. 

Expanding beyond prompt engineering, methods such as LaCLIP~\cite{laclip}, LaBo~\cite{labo}, and VFC~\cite{vfc} refine CLIP’s visual-textual alignment by leveraging LLM-enriched captions to improve performance across diverse tasks and domains. Similarly, ARGUE~\cite{attr3} employs LLM-generated attributes, followed by attribute sampling, to enhance visual-semantic mapping. Another perspective is introduced by Kim \etal~\cite{kim2024aapl}, which integrates visual attribute learning into prompts using contrastive learning.

Despite these advancements, the nuanced interplay between LLM-generated attributes and visual data remains underexplored. This is where one of the novelties of \textsc{OSLoPrompt} lies, bridging this gap by effectively integrating LLM-driven semantic attributes with visual cues to enhance open-set recognition and domain generalization in the low-supervision setting.

\section{Class attributes, and the pseudo-open class names generated by GPT-4o}

In Table \ref{tab:attribute} and Table \ref{tab:fine_grained_samples} highlight the class-wise four attributes in $\mathcal{A}$ and the pseudo-open class names generated in $\mathcal{C}^{\text{open}}$, both using GPT-4o.
\begin{table*}[h]
\centering
\renewcommand{\arraystretch}{1.3}
\setlength{\tabcolsep}{12pt}
\small
\caption{Closed-set classes and their attributes generated by GPT-4o for the PACS dataset\cite{pacs}.}
\label{tab:attribute}
\vspace{-3mm}
\begin{adjustbox}{width=0.4\linewidth}
\begin{tabular}{@{}cc@{}}
\toprule
\rowcolor[HTML]{EFEFEF} \textbf{Class} & \textbf{Attributes} \\ 
\midrule
Dog      & Fur, snout, tail, paw pads          \\ 
Elephant & Large ears, trunk, tusks, wrinkled skin \\ 
Giraffe  & Long neck, spotted pattern, horns, slender legs \\ 
Guitar   & Curved body, strings, fretboard, soundhole \\ 
Horse    & Mane, hooves, muscular build, tail \\ 
House    & Roof, windows, doors, chimney \\ 
\bottomrule
\end{tabular}
\end{adjustbox}
\vspace{-2mm}
\end{table*}

\begin{table*}[htbp]
\centering
\caption{Fine-grained pseudo-open-set class names vs. closed-set class names for the PACS dataset\cite{pacs} generated by GPT-4o.}
\label{tab:fine_grained_samples}
\vspace{-3mm}
\begin{adjustbox}{width=0.9\textwidth}
\begin{tabular}{@{}p{0.25\textwidth}|p{0.7\textwidth}@{}}
\toprule
\rowcolor[HTML]{EFEFEF} \textbf{Closed-set Classes} & \textbf{Related Fine-Grained Pseudo Open-Set Classes Outputted by GPT-4o} \\ 
\midrule
\textbf{Dog}      & Wolf, Fox, Coyote, Jackal, Dhole, Fennec Fox, Hyena, Maned Wolf \\ 
\textbf{Elephant} & Mastodon, Woolly Mammoth, Rhinoceros, Hippopotamus \\ 
\textbf{Giraffe}  & Okapi, Pronghorn, Impala, Sable Antelope, Kudu, Eland, Gazelle, Springbok, Nyala, Gerenuk \\ 
\textbf{Guitar}   & Mandolin, Banjo, Lute, Bouzouki, Sitar, Balalaika, Charango, Oud, Lyre, Zither \\ 
\textbf{Horse}    & Zebra, Donkey, Onager, Kiang, Tarpan, Wild Ass, Quagga \\ 
\textbf{House}    & Castle, Hut, Palace \\ 
\midrule
\multicolumn{2}{@{}p{\textwidth}}{\textbf{Additional Fine-Grained Pseudo Open-Set Classes:}} \\ 
\multicolumn{2}{@{}p{\textwidth}}{Alpaca, Emu, Lynx, Peacock, Ferret, Armadillo, Pangolin, Tamarin, Mongoose, Marten, Caracal, Serval, Ocelot, Civet, Quokka, Wallaby, Pademelon, Koala, Pika, Aye-aye, Tarsier, Wombat, Kinkajou, Agouti, Coati, Cuscus, Galago, Jerboa, Marmoset} \\
\bottomrule
\end{tabular}
\end{adjustbox}
\vspace{-2mm}
\end{table*}

\section{Implementation details for both comparative and ablation methods}
\label{sec:implementation}
We evaluate our proposed methods against several state-of-the-art approaches using their official implementations, incorporating necessary modifications to ensure compatibility with 1-shot and 5-shot settings. For \textbf{PromptSRC} \cite{promptsrc} and \textbf{\textsc{StyLIP}} \cite{stylip}, we extend their frameworks by integrating synthetic open samples as outlined in our method and employing an "unknown" prompt. These enhancements refine the original designs to more effectively address open-set scenarios. Similarly, for \textbf{CLIP+OpenMax} \cite{osr1}, we adapt the approach by computing Mean Activation Vectors (MAVs) for each class using a modified data loader and optimizing thresholds for improved open-set recognition accuracy.

For \textbf{MORGAN} \cite{morgan} and \textbf{2LM} \cite{2lm}, we implement meta-learning strategies on the dataset $\mathcal{D}$, modifying the backbone for \textbf{MORGAN} and incorporating OpenMax for open-set recognition in \textbf{2LM}, analogous to \textbf{CLIP+OpenMax}. Meta-training is conducted over 30 episodes for both methods, during which convergence was observed. Methods such as \textbf{\textsc{StyLIP}} \cite{stylip} and \textbf{ODG-Net} \cite{odg-net} are evaluated using the official implementations provided by their authors. All methods are trained for the default number of epochs specified in their respective implementations.

For the ImageNet experiments, we extend \textbf{\textsc{OSLoPrompt}} by introducing additional domain-specific prompts selected from ImageNet templates provided in the official CLIP implementation \cite{clip}. To ensure consistency, we use a ViT-B/32 backbone across all models. Optimization is performed using the SGD optimizer with a learning rate of 0.0035 over six epochs. These adjustments ensure robust and fair comparisons across methods, emphasizing adaptability in joint open-set and low-shot domain generalization (DG) scenarios. All comparative methods are evaluated in the LSOSDG setting for consistency.

In the ablation experiments, we generate pseudo-open samples using a Mix-up-based \cite{cumix} approach, with $\lambda$ uniformly sampled from $[0.3, 0.7]$. For two samples, $x_i^s \in \mathcal{X}_s$ and $x_j^{s'} \in \mathcal{X}_{s'}$, from source domains $s$ and $s'$, respectively, the generated pseudo-open sample $x^{\text{open}} \in \mathcal{D}^{\text{open}}$ is defined as:
\[
x^{\text{open}} = \lambda x_i^s + (1 - \lambda) x_j^{s'}
\]
where $\lambda \in [0.3, 0.7]$. For manual domain-specific prompting, ad hoc attributes are created by concatenating the four attributes of each class along with the class name. For instance, the attributes of the dog class, as shown in Table \ref{tab:attribute}, are "Fur, snout, tail, paw pad." The resulting manual prompt for the dog class becomes: "A \{domain\} of dog with Fur, snout, tail, paw pad."

For the image-conditioning experiments, projected image features from the Meta-Net are incorporated into the prompts before being passed to the CLIP \cite{clip} text encoder. The Meta-Net consists of two linear layers with an intermediate representation size of 32, with a ReLU activation \cite{relu} applied between the layers. Additionally, we evaluate two types of domain-generic prompting: (1) textual prompting and (2) textual prompting combined with image conditioning, inspired by \textbf{CoOp} \cite{coop} and \textbf{CoCoOp} \cite{cocoop}, respectively. In both cases, the context length is fixed at 4.

\begin{figure*}[!h]
    \centering
    \caption{Pseudo open images generated by \cite{odgclip} and \textsc{OSLoPrompt}, given the known-class images, for PACS.}
    \includegraphics[width=\textwidth]{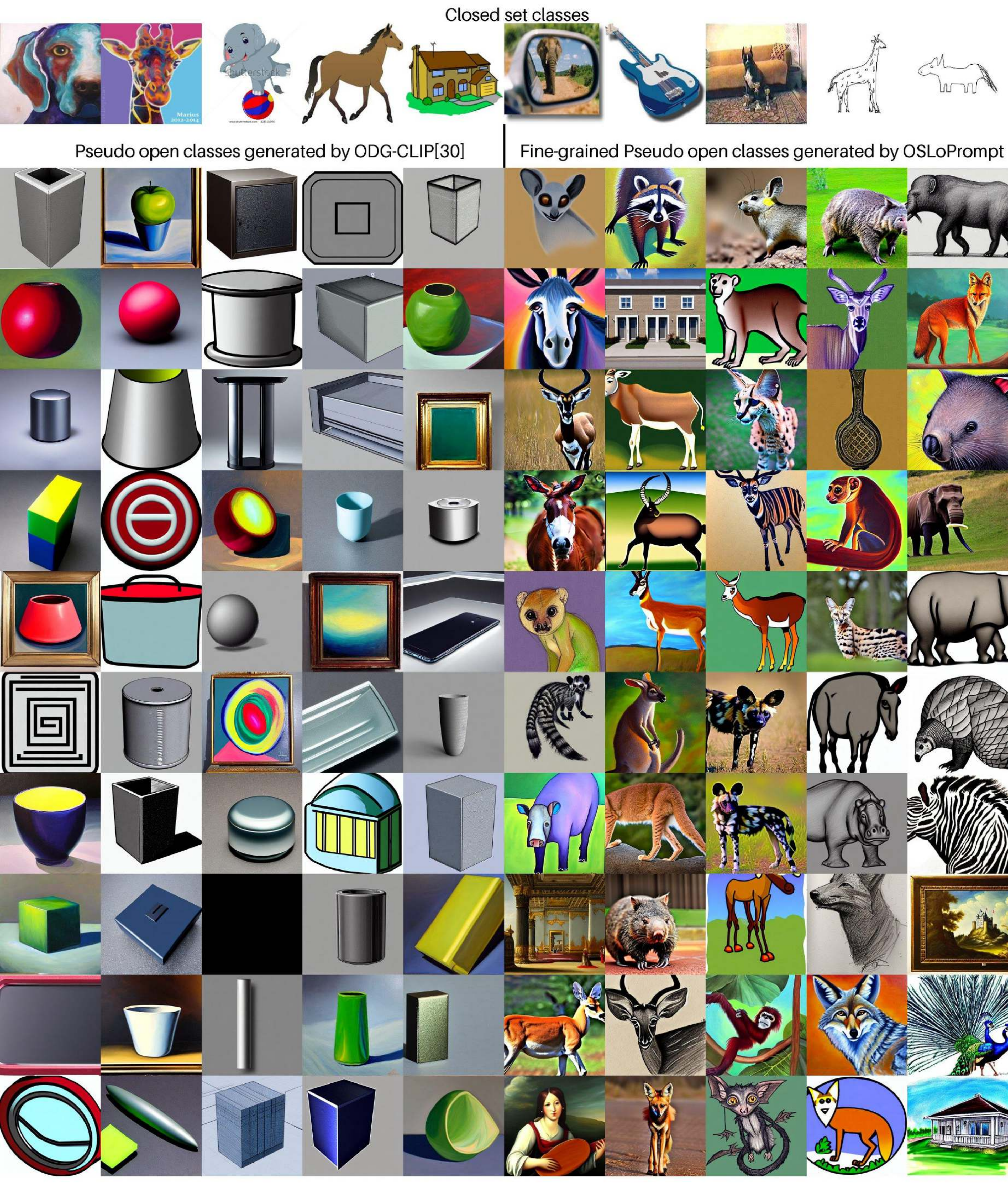}  
    \label{fig:grid}
\end{figure*}

\section{Generated pseudo-open samples by ODG-CLIP \cite{odgclip} and \textsc{OSLoPrompt}}\label{sec:vis}
Fig. \ref{fig:grid} compares the pseudo-open images generated by the method of ODG-CLIP, which are mostly coarse-grained, and \textsc{OSLoPrompt}, which are fine-grained in nature, given the closed-set images. In Table \ref{tab:fid}, we compare the FID distance \cite{dowson1982frechet} between the closed set images and the generated pseudo-open samples. The proposed fine-grained pseudo-open samples are highly similar to closed-set samples when compared to ODG-CLIP's synthesized pseudo-open samples. The FID score in our case is low by 4.9 points than that of ODG-CLIP.
\begin{table}[h]
\centering
\caption{FID between the closed-set images and generated pseudo-open images for the PACS dataset.}
\label{tab:fid}
\vspace{-3mm}
\begin{adjustbox}{width=0.4\linewidth}
\begin{tabular}{@{}lc@{}}
\toprule
\rowcolor[HTML]{EFEFEF} \textbf{Pseudo-open image synthesis method} & \textbf{FID score} \\ 
\midrule
ODG-CLIP\cite{odgclip} pseudo open samples & 13.04 \\ 
Fine-grained pseudo open samples (ours)    & \textbf{8.14} \\ 
\bottomrule
\end{tabular}
\end{adjustbox}
\vspace{-2mm}
\end{table}

\section{Detailed results on all the datasets}

We report the detailed results for all the domain combinations for the five datasets in Table \ref{tab:minidomainnet_domain_performance} - \ref{table:pacs_domain_performance}. In the leave-one-domain-out protocol, all but one domains are considered as sources, while the rest acts as the target domain.

\begin{table*}[!h]
\centering
\renewcommand{\arraystretch}{1.2}
\setlength{\tabcolsep}{5pt}
\caption{Accuracy over different target domains in the Mini-DomainNet dataset. The 1-shot results are at the top, and the 5-shot results are at the bottom. For each case, the other domains are considered as the source domains.}
\begin{tabular}{lcccccccccc}
\toprule
\multirow{2}{*}{\textbf{Method}} & \multicolumn{2}{c}{\textbf{Clipart}} & \multicolumn{2}{c}{\textbf{Painting}} & \multicolumn{2}{c}{\textbf{Sketch}} & \multicolumn{2}{c}{\textbf{Real}} & \multirow{2}{*}{\textbf{Avg Acc}} & \multirow{2}{*}{\textbf{Avg H-score}} \\
\cmidrule(lr){2-3} \cmidrule(lr){4-5} \cmidrule(lr){6-7} \cmidrule(lr){8-9}
 & \textbf{Acc} & \textbf{H-score} & \textbf{Acc} & \textbf{H-score} & \textbf{Acc} & \textbf{H-score} & \textbf{Acc} & \textbf{H-score} & & \\
\midrule
CLIP + OpenMax \cite{osr1} & 20.50 & 14.28 & 19.50 & 32.42 & 7.69 & 32.42 & 20.00 & 33.08 & 16.92 & 28.05 \\
CLIPN \cite{clipn} & 47.43 & 40.68 & 50.81 & 39.56 & 42.60 & 39.91 & 49.67 & 43.50 & 47.63 & 40.91 \\
MORGAN \cite{morgan} & 26.59 & 11.28 & 14.16 & 16.95 & 28.41 & 23.34 & 20.44 & 11.25 & 22.40 & 15.70 \\
STYLIP \cite{stylip} & 62.48 & 59.45 & \cellcolor{red!20}60.78 & 56.15 & 47.24 & 43.72 & \cellcolor{red!20}67.25 & \cellcolor{red!20}70.51 & 59.44 & 57.46 \\
PROMPTSRC \cite{promptsrc} & 27.38 & 19.71 & 26.91 & 25.54 & 20.24 & 21.72 & 26.25 & 14.79 & 25.20 & 20.44 \\
2LM \cite{2lm} & 28.06 & 18.71 & 23.94 & 19.62 & 21.36 & 20.88 & 24.65 & 11.78 & 24.50 & 17.75 \\
ODG-Net \cite{odg-net} & 21.76 & 17.50 & 24.42 & 27.99 & 21.98 & 11.44 & 20.04 & 19.40 & 22.05 & 19.08 \\
MEDIC \cite{medic} & 26.53 & 21.99 & 24.63 & 24.08 & 19.84 & 20.67 & 23.93 & 9.48 & 23.73 & 19.05 \\
SCI-PD \cite{scipd} & 17.50 & 22.95 & 15.50 & 23.20 & 16.00 & 23.12 & 16.00 & 24.04 & 16.25 & 23.33 \\
ODG-CLIP \cite{odgclip} & \cellcolor{red!20}66.84 & \textbf{76.10} & 54.74 & \cellcolor{red!20}66.25 & \cellcolor{red!20}59.47 & \textbf{60.55} & 63.16 & 59.12 & \cellcolor{red!20}61.05 & \cellcolor{red!20}65.50 \\
OSLOPROMPT & \textbf{76.00} & \cellcolor{red!20}66.00 &\textbf{ 61.50} & \textbf{66.43} & \textbf{60.00} & \cellcolor{red!20}59.15 & \textbf{78.50 }& \textbf{78.69} & \textbf{69.00} & \textbf{67.57} \\
\midrule
CLIP + OpenMax \cite{osr1} & 31.50 & 47.17 & 30.00 & 45.54 & 32.82 & 48.93 & 35.50 & 51.17 & 32.46 & 48.20 \\
CLIPN \cite{clipn} & 53.39 & 44.65 & 58.53 & 53.04 & 55.50 & 50.91 & 55.71 & 46.31 & 55.78 & 48.53 \\
MORGAN \cite{morgan} & 32.10 & 21.15 & 36.26 & 29.31 & 35.26 & 32.32 & 47.62 & 25.45 & 37.81 & 27.06 \\
STYLIP \cite{stylip} & 64.12 & \cellcolor{red!20}60.03 & \cellcolor{red!20}62.52 & \cellcolor{red!20}60.65 & 53.87 & 44.12 & 75.61 & 77.93 & 64.03 & 60.68 \\
PROMPTSRC \cite{promptsrc} & 37.68 & 32.38 & 33.21 & 32.00 & 39.80 & 31.04 & 34.80 & 29.86 & 36.37 & 31.32 \\
2LM \cite{2lm} & 38.55 & 28.27 & 34.34 & 25.47 & 35.19 & 35.10 & 45.65 & 25.94 & 38.43 & 28.70 \\
ODG-Net \cite{odg-net} & 45.69 & 30.53 & 38.53 & 20.86 & 39.71 & 23.41 & 35.88 & 20.08 & 39.95 & 23.72 \\
MEDIC \cite{medic} & 39.87 & 33.08 & 30.42 & 28.23 & 35.83 & 32.16 & 41.68 & 28.94 & 36.95 & 30.60 \\
SCI-PD \cite{scipd} & 19.50 & 28.23 & 23.50 & 34.60 & 15.50 & 24.18 & 26.50 & 35.28 & 21.25 & 30.57 \\
ODG-CLIP \cite{odgclip} & \cellcolor{red!20}79.00 & 52.12 & 60.53 & 58.54 & \textbf{78.97} & \textbf{88.03} & \cellcolor{red!20}79.00 & \cellcolor{red!20}63.28 & \cellcolor{red!20}74.38 & \cellcolor{red!20}65.49 \\
OSLOPROMPT & \textbf{86.00} & \textbf{63.68} & \textbf{63.00} & \textbf{66.06} & \cellcolor{red!20}63.58 & \cellcolor{red!20}62.03 & \textbf{85.50} & \textbf{74.53} & \textbf{74.52} & \textbf{66.58} \\
\bottomrule
\end{tabular}
\label{tab:minidomainnet_domain_performance}
\end{table*}

\begin{table*}[!h]
\centering
\renewcommand{\arraystretch}{1.2}
\setlength{\tabcolsep}{5pt}
\caption{Accuracy over different target domains in the Multi-dataset benchmark. The 1-shot results are shown at the top and the 5-shot results at the bottom. For each case, the other domains are considered as the source domains.}
\begin{tabular}{lcccccccccc}
\toprule
\multirow{2}{*}{\textbf{Method}} & \multicolumn{2}{c}{\textbf{Clipart}} & \multicolumn{2}{c}{\textbf{Painting}} & \multicolumn{2}{c}{\textbf{Sketch}} & \multicolumn{2}{c}{\textbf{Real}} & \multirow{2}{*}{\textbf{Avg Acc}} & \multirow{2}{*}{\textbf{Avg H-score}} \\
\cmidrule(lr){2-3} \cmidrule(lr){4-5} \cmidrule(lr){6-7} \cmidrule(lr){8-9}
 & \textbf{Acc} & \textbf{H-score} & \textbf{Acc} & \textbf{H-score} & \textbf{Acc} & \textbf{H-score} & \textbf{Acc} & \textbf{H-score} & & \\
\midrule
CLIP + OpenMax \cite{osr1} & 10.00 & 18.17 & 16.15 & 27.80 & 12.79 & 22.67 & 8.02 & 14.85 & 11.74 & 20.87 \\
CLIPN \cite{clipn} & 41.51 & 37.31 & 34.53 & 34.48 & 36.38 & 34.49 & 46.92 & 38.82 & 39.84 & 36.28 \\
MORGAN \cite{morgan} & 31.66 & 46.72 & 26.76 & 36.43 & 23.48 & 31.91 & 38.08 & 33.96 & 30.00 & 37.26 \\
STYLIP \cite{stylip} & 56.32 & 50.22 & 49.10 & 45.16 & 39.92 & 35.28 & 60.65 & 59.89 & 51.50 & 47.64 \\
PROMPTSRC \cite{promptsrc} & 27.38 & 25.06 & 30.51 & 31.82 & 29.10 & 31.97 & 33.65 & 35.85 & 30.16 & 31.18 \\
2LM \cite{2lm} & 28.24 & 38.49 & 27.47 & 32.13 & 31.19 & 32.19 & 32.00 & 36.37 & 29.73 & 34.80 \\
ODG-Net \cite{odg-net} & 25.88 & 25.48 & 24.05 & 35.85 & 25.80 & 27.46 & 40.92 & 28.79 & 29.16 & 29.40 \\
MEDIC \cite{medic} & 31.62 & 34.15 & 26.31 & 30.99 & 27.71 & 31.29 & 35.75 & 36.00 & 30.35 & 33.11 \\
SCI-PD \cite{scipd} & 15.37 & 11.78 & 17.62 & 21.48 & 20.52 & 24.88 & 14.30 & 18.56 & 16.95 & 19.18 \\
ODG-CLIP \cite{odgclip} & \cellcolor{red!20}66.42 & \cellcolor{red!20}71.71 & \cellcolor{red!20}55.25 & \cellcolor{red!20}65.09 & \cellcolor{red!20}58.21 & \textbf{65.71} & \cellcolor{red!20}75.08 & \cellcolor{red!20}75.60 & \cellcolor{red!20}63.74 & \cellcolor{red!20}69.53 \\
OSLOPROMPT & \textbf{79.04} & \textbf{78.18} & \textbf{66.06} & \textbf{69.49} &\textbf{ 72.55} & \cellcolor{red!20}62.74 & \textbf{87.55} & \textbf{87.55} &\textbf{ 76.30} & \textbf{74.49} \\
\midrule
CLIP + OpenMax \cite{osr1} & 48.04 & 63.09 & 62.63 & 73.06 & 51.72 & 65.22 & 63.97 & 73.97 & 56.59 & 68.84 \\
CLIPN \cite{clipn} & 42.28 & 38.18 & 44.76 & 39.56 & 47.25 & 36.32 & 51.69 & 42.86 & 46.50 & 39.23 \\
MORGAN \cite{morgan} & 35.65 & 45.48 & 38.28 & 44.86 & 30.48 & 34.62 & 37.48 & 46.26 & 35.47 & 42.80 \\
STYLIP \cite{stylip} & 59.70 & 55.58 & 52.05 & 46.25 & 45.13 & 37.00 & 61.43 & 60.21 & 54.58 & 49.76 \\
PROMPTSRC \cite{promptsrc} & 34.44 & 36.36 & 31.50 & 35.52 & 38.40 & 39.75 & 38.39 & 40.86 & 35.68 & 38.12 \\
2LM \cite{2lm} & 35.31 & 37.22 & 34.74 & 34.43 & 31.68 & 31.15 & 38.44 & 38.72 & 35.04 & 35.38 \\
ODG-Net \cite{odg-net} & 29.62 & 32.93 & 28.59 & 33.63 & 32.41 & 39.75 & 46.16 & 41.41 & 34.20 & 36.93 \\
MEDIC \cite{medic} & 32.24 & 34.92 & 34.05 & 36.31 & 36.00 & 34.05 & 39.38 & 39.74 & 35.42 & 36.26 \\
SCI-PD \cite{scipd} & 27.58 & 28.05 & 31.26 & 22.56 & 39.41 & 38.49 & 30.54 & 26.46 & 32.20 & 28.89 \\
ODG-CLIP \cite{odgclip} & \cellcolor{red!20}67.65 & \cellcolor{red!20}73.33 & \cellcolor{red!20}71.63 & \textbf{76.26} & \textbf{75.60 }& \cellcolor{red!20}71.52 & \cellcolor{red!20}82.71 & \cellcolor{red!20}83.46 & \cellcolor{red!20}74.40 & \cellcolor{red!20}76.14 \\
OSLOPROMPT & \textbf{83.21} & \textbf{78.69} & \textbf{73.12} & \cellcolor{red!20}75.36 & \cellcolor{red!20}71.86 & \textbf{76.13} & \textbf{90.81} & \textbf{90.00} & \textbf{79.75} & \textbf{80.05} \\
\bottomrule
\end{tabular}
\label{tab:multidataset_domain_performance}
\end{table*}

\begin{table*}[!h]
\centering
\renewcommand{\arraystretch}{1.2}
\setlength{\tabcolsep}{5pt}
\caption{Accuracy and H-score over different target domains in the Office-Home dataset. The 1-shot results are shown at the top and the 5-shot results at the bottom. For each case, the other domains are considered as the source domains.}
\begin{tabular}{lcccccccccc}
\toprule
\multirow{2}{*}{\textbf{Method}} & \multicolumn{2}{c}{\textbf{Clipart}} & \multicolumn{2}{c}{\textbf{Product}} & \multicolumn{2}{c}{\textbf{Real World}} & \multicolumn{2}{c}{\textbf{Art}} & \multirow{2}{*}{\textbf{Avg Acc}} & \multirow{2}{*}{\textbf{Avg H-score}} \\
\cmidrule(lr){2-3} \cmidrule(lr){4-5} \cmidrule(lr){6-7} \cmidrule(lr){8-9}
 & \textbf{Acc} & \textbf{H-score} & \textbf{Acc} & \textbf{H-score} & \textbf{Acc} & \textbf{H-score} & \textbf{Acc} & \textbf{H-score} & & \\
\midrule
CLIP + OpenMax \cite{osr1} & 15.91 & 31.45 & 18.70 & 31.45 & 25.62 & 40.58 & 19.74 & 27.09 & 19.99 & 32.64 \\
CLIPN \cite{clipn} & \cellcolor{red!20}48.52 & 36.07 & 40.41 & 31.50 & 44.50 & 31.70 & 43.27 & 32.06 & 44.18 & 32.83 \\
MORGAN \cite{morgan} & 24.88 & 16.73 & 4.59 & 14.95 & 30.28 & 24.17 & 17.07 & 18.19 & 19.21 & 18.51 \\
STYLIP \cite{stylip} & 42.35 & 20.37 & \cellcolor{red!20}60.00 & 15.79 & \cellcolor{red!20}62.51 & 34.13 & \cellcolor{red!20}44.49 & 17.52 & \cellcolor{red!20}52.34 & 21.95 \\
PROMPTSRC \cite{promptsrc} & 22.37 & 17.44 & 14.30 & 12.93 & 30.10 & 14.10 & 21.30 & 14.94 & 22.02 & 14.85 \\
2LM \cite{2lm} & 27.88 & 13.47 & 10.12 & 14.95 & 30.28 & 13.61 & 16.93 & 12.35 & 21.30 & 13.60 \\
ODG-Net \cite{odg-net} & 31.30 & 11.44 & 1.89 & 10.89 & 26.21 & 13.27 & 22.48 & 10.18 & 20.47 & 11.45 \\
MEDIC \cite{medic} & 26.70 & 11.25 & 10.83 & 11.29 & 28.68 & 11.64 & 19.03 & 12.80 & 21.31 & 11.75 \\
SCI-PD \cite{scipd} & 25.00 & 34.51 & 29.20 & 38.44 & 48.76 & \cellcolor{red!20}56.48 & 38.10 & 47.79 & 35.27 & 44.31 \\
ODG-CLIP \cite{odgclip} & 46.21 & \cellcolor{red!20}54.72 & 50.41 & \cellcolor{red!20}55.30 & 58.68 & 46.87 & 39.47 & \cellcolor{red!20}54.83 & 48.69 & \cellcolor{red!20}52.93 \\
OSLOPROMPT & \textbf{59.09} & \textbf{63.38} & \textbf{81.30} &\textbf{ 67.35} & \textbf{79.33} & \textbf{70.48} & \textbf{59.21} & \textbf{54.96} & \textbf{69.73} & \textbf{64.04} \\
\midrule
CLIP + OpenMax \cite{osr1} & 29.55 & 44.32 & 29.61 & 44.39 & 30.58 & 42.76 & 52.63 & \cellcolor{red!20}65.64 & 35.59 & 49.28 \\
CLIPN \cite{clipn} & 44.22 & 40.51 & 46.81 & 40.46 & 49.38 & 40.47 & 51.34 & 39.87 & 47.94 & 40.33 \\
MORGAN \cite{morgan} & 40.07 & 17.62 & 30.93 & 11.12 & 37.63 & 29.23 & 36.17 & 16.55 & 36.20 & 18.63 \\
STYLIP \cite{stylip} & \cellcolor{red!20}50.62 & 42.91 & 65.32 & 40.28 & 78.42 & 51.98 & \cellcolor{red!20}53.11 & 34.67 & \cellcolor{red!20}61.87 & 42.46 \\
PROMPTSRC \cite{promptsrc} & 32.30 & 20.82 & 30.20 & 22.08 & 28.21 & 16.98 & 33.70 & 21.51 & 31.10 & 20.35 \\
2LM \cite{2lm} & 35.97 & 11.23 & 28.33 & 14.91 & 28.01 & 25.35 & 25.22 & 24.32 & 29.38 & 18.95 \\
ODG-Net \cite{odg-net} & 43.65 & 22.11 & 36.01 & 12.59 & 28.74 & 16.05 & 29.68 & 13.08 & 34.52 & 15.96 \\
MEDIC \cite{medic} & 35.27 & 16.11 & 27.42 & 16.31 & 31.66 & 20.59 & 27.24 & 20.77 & 30.40 & 18.45 \\
SCI-PD \cite{scipd} & 25.76 & 34.97 & 35.77 & 41.26 & 45.40 & \textbf{57.40} & 25.00 & 36.36 & 32.98 & 42.50 \\
ODG-CLIP \cite{odgclip} & 45.45 & \cellcolor{red!20}47.90 & \cellcolor{red!20}73.17 & \textbf{81.22} & \cellcolor{red!20}71.07 & 30.61 & 31.57 & 37.50 & 55.32 & \cellcolor{red!20}49.31 \\
OSLOPROMPT & \textbf{61.36} & \textbf{57.68} & \textbf{91.86 }& \cellcolor{red!20}66.81 & \textbf{80.99} & \cellcolor{red!20}57.03 & \textbf{67.10} & \textbf{66.80}& \textbf{75.33} & \textbf{62.08 }\\
\bottomrule
\end{tabular}
\label{tab:officehome_domain_performance}
\end{table*}

\begin{table*}[!h]
\centering
\renewcommand{\arraystretch}{1.2}
\setlength{\tabcolsep}{5pt}
\caption{Accuracy and H-score across target domains in the VLCS dataset. The 1-shot results are shown at the top and the 5-shot results at the bottom. For each case, the other domains are considered as the source domains.}
\begin{tabular}{lcccccccccc}
\toprule
\multirow{2}{*}{\textbf{Method}} & \multicolumn{2}{c}{\textbf{CALTECH}} & \multicolumn{2}{c}{\textbf{SUN09}} & \multicolumn{2}{c}{\textbf{VOC2007}} & \multicolumn{2}{c}{\textbf{LABELME}} & \textbf{Avg Acc} & \textbf{Avg H-score} \\
\cmidrule(lr){2-3} \cmidrule(lr){4-5} \cmidrule(lr){6-7} \cmidrule(lr){8-9}
 & \textbf{Acc} & \textbf{H-score} & \textbf{Acc} & \textbf{H-score} & \textbf{Acc} & \textbf{H-score} & \textbf{Acc} & \textbf{H-score} & & \\
\midrule
CLIP + OpenMax \cite{osr1} & 22.94 & 37.32 & 2.08 & 4.08 & 21.42 & 34.85 & 35.92 & 51.08 & 20.59 & 31.83 \\
CLIPN \cite{clipn} & 25.70 & 19.14 & 21.43 & 19.16 & 27.40 & 18.98 & 26.83 & 20.66 & 25.34 & 19.49 \\
MORGAN \cite{morgan} & 32.40 & 25.58 & 22.81 & 31.12 & 39.77 & 28.26 & 30.44 & 23.91 & 31.35 & 27.22 \\
STYLIP \cite{stylip} & 11.04 & 19.23 & 21.45 & 31.17 & 25.75 & 35.54 & \cellcolor{red!20}53.52 & \cellcolor{red!20}52.48 & 27.94 & 34.61 \\
PROMPTSRC \cite{promptsrc} & 26.28 & 22.25 & 29.58 & 27.59 & 19.34 & 12.07 & 24.71 & 18.26 & 24.98 & 20.04 \\
2LM \cite{2lm} & 33.09 & 29.65 & 22.70 & 32.38 & 37.37 & 27.01 & 33.28 & 26.01 & 31.61 & 28.76 \\
ODG-Net \cite{odg-net} & 34.93 & 27.62 & 21.76 & 33.37 & 41.08 & 30.25 & 31.57 & 25.42 & 32.33 & 29.17 \\
MEDIC \cite{medic} & 33.41 & 27.37 & 27.86 & 27.08 & 35.62 & 23.80 & 34.88 & 26.86 & 32.94 & 26.28 \\
SCI-PD \cite{scipd} & 22.63 & 23.03 & 26.42 & 25.77 & 11.90 & 12.13 & 18.56 & 17.47 & 19.88 & 19.60 \\
ODG-CLIP \cite{odgclip} &\cellcolor{red!20}80.62 & \cellcolor{red!20}87.25 & \cellcolor{red!20}54.10 & \cellcolor{red!20}50.11 & \cellcolor{red!20}52.95 & \cellcolor{red!20}50.61 & 22.05 & 30.84 & \cellcolor{red!20}52.43 & \cellcolor{red!20}54.70 \\
OSLOPROMPT & \textbf{99.47} & \textbf{99.73} & \textbf{62.75} & \textbf{66.60} &\textbf{ 78.02} & \textbf{79.55} & \textbf{75.30} & \textbf{61.69}& \textbf{78.89} & \textbf{76.89} \\
\midrule
CLIP + OpenMax \cite{osr1} & \cellcolor{red!20}77.98 & \cellcolor{red!20}87.63 & 55.35 & \cellcolor{red!20}67.57 & \cellcolor{red!20}64.41 & \cellcolor{red!20}75.17 & \cellcolor{red!20}67.25 & \cellcolor{red!20}68.60 & \cellcolor{red!20}66.25 & \cellcolor{red!20}74.74 \\
CLIPN \cite{clipn} & 27.33 & 27.07 & 35.46 & 29.75 & 36.52 & 26.99 & 32.38 & 28.00 & 32.92 & 27.95 \\
MORGAN \cite{morgan} & 34.59 & 40.39 & 48.41 & 39.91 & 39.39 & 39.50 & 46.27 & 35.01 & 42.16 & 38.70 \\
STYLIP \cite{stylip} & 46.01 & 55.80 & 41.25 & 45.47 & 42.09 & 49.02 & 53.75 & 45.44 & 45.78 & 48.93 \\
PROMPTSRC \cite{promptsrc} & 36.95 & 29.57 & 35.18 & 32.19 & 34.27 & 33.91 & 38.24 & 33.77 & 36.16 & 32.36 \\
2LM \cite{2lm} & 37.22 & 38.42 & 44.73 & 37.52 & 38.03 & 36.79 & 46.71 & 36.69 & 41.67 & 37.36 \\
ODG-Net \cite{odg-net} & 36.33 & 40.18 & 52.73 & 38.70 & 38.95 & 35.51 & 44.20 & 36.46 & 43.05 & 37.71 \\
MEDIC \cite{medic} & 35.53 & 36.20 & 44.33 & 36.31 & 40.17 & 35.95 & 42.09 & 34.76 & 40.53 & 35.56 \\
SCI-PD \cite{scipd} & 30.24 & 30.77 & 32.71 & 34.01 & 29.16 & 28.29 & 28.34 & 28.84 & 30.11 & 30.48 \\
ODG-CLIP \cite{odgclip} & 76.96 & 86.86 & \textbf{70.82} & 42.43 & 47.56 & 48.22 & 56.39 & 50.06 & 62.93 & 56.89 \\
OSLOPROMPT & \textbf{98.95} & \textbf{99.39} & \cellcolor{red!20}64.02 & \textbf{68.26 }& \textbf{76.86} & \textbf{76.86} & \textbf{76.33} & \textbf{64.84} & \textbf{79.04} & \textbf{77.34} \\
\bottomrule
\end{tabular}
\label{table:vlcs_domain_performance}
\end{table*}

\begin{table*}[!h]
\centering
\renewcommand{\arraystretch}{1.2}
\setlength{\tabcolsep}{5pt}
\caption{Accuracy and H-score across target domains in the PACS dataset. The 1-shot results are shown at the top and the 5-shot results are at the bottom. For each case, the other domains are considered as the source domains.}
\begin{tabular}{lcccccccccc}
\toprule
\multirow{2}{*}{\textbf{Method}} & \multicolumn{2}{c}{\textbf{Art Painting}} & \multicolumn{2}{c}{\textbf{Photo}} & \multicolumn{2}{c}{\textbf{Sketch}} & \multicolumn{2}{c}{\textbf{Cartoon}} & \textbf{Avg Acc} & \textbf{Avg H-score} \\
\cmidrule(lr){2-3} \cmidrule(lr){4-5} \cmidrule(lr){6-7} \cmidrule(lr){8-9}
 & \textbf{Acc} & \textbf{H-score} & \textbf{Acc} & \textbf{H-score} & \textbf{Acc} & \textbf{H-score} & \textbf{Acc} & \textbf{H-score} & & \\
\midrule
CLIP + OpenMax \cite{osr1} & 33.83 & 49.78 & 14.70 & 25.63 & 5.00 & 9.52 & 27.44 & 42.95 & 20.24 & 31.97 \\
CLIPN \cite{clipn} & 63.34 & 54.15 & 65.46 & 53.62 & 63.80 & 56.84 & 63.50 & 58.55 & 64.03 & 55.79 \\
MORGAN \cite{morgan} & 28.92 & 0.94 & 44.35 & 21.44 & 32.22 & 31.96 & 44.13 & 21.91 & 37.40 & 19.06 \\
STYLIP \cite{stylip} & \cellcolor{red!20}73.17 & \cellcolor{red!20}78.96 & \cellcolor{red!20}73.10 & \cellcolor{red!20}77.68 & 73.49 & 43.85 & 79.78 & 43.47 & \cellcolor{red!20}74.89 & 60.99 \\
PROMPTSRC \cite{promptsrc} & 30.15 & 17.47 & 35.83 & 30.98 & 37.42 & 33.05 & 39.47 & 26.84 & 35.72 & 27.09 \\
2LM \cite{2lm} & 32.53 & 9.57 & 36.37 & 24.89 & 31.54 & 30.88 & 40.45 & 20.35 & 35.22 & 21.42 \\
ODG-Net \cite{odg-net} & 36.67 & 9.00 & 29.73 & 12.49 & 32.56 & 35.56 & 40.32 & 29.61 & 34.82 & 21.67 \\
MEDIC \cite{medic} & 29.47 & 11.12 & 34.33 & 20.61 & 35.54 & 28.84 & 36.28 & 25.03 & 33.91 & 21.40 \\
SCI-PD \cite{scipd} & 26.30 & 28.75 & 19.50 & 22.07 & 22.31 & 24.88 & 25.47 & 27.65 & 23.40 & 25.84 \\
ODG-CLIP \cite{odgclip} & 51.34 & 62.53 & 59.05 & 73.28 & \cellcolor{red!20}82.30 & \cellcolor{red!20}87.93 & \cellcolor{red!20}82.88 & \cellcolor{red!20}78.51 & 68.89 & \cellcolor{red!20}75.56 \\
OSLOPROMPT & \textbf{91.61} & \textbf{93.43} & \textbf{99.43} & \textbf{99.71} & \textbf{82.75} & \textbf{92.70} & \textbf{97.06} & \textbf{93.59} & \textbf{92.71} & \textbf{94.86} \\
\midrule
CLIP + OpenMax \cite{osr1} & 63.79 & 77.55 & 74.56 & 85.26 & 60.47 & 74.83 & 76.17 & 86.29 & 68.75 & 80.98 \\
CLIPN \cite{clipn} & 78.10 & 69.36 & 78.27 & 71.89 & \cellcolor{red!20}77.41 & 71.88 & 78.36 & 71.42 & 78.04 & 71.14 \\
MORGAN \cite{morgan} & 50.39 & 33.03 & 38.52 & 29.40 & 45.84 & 8.95 & 50.34 & 24.86 & 46.27 & 24.06 \\
STYLIP \cite{stylip} & 75.24 & 79.67 & 87.26 & 88.31 & 74.45 & 50.78 & 83.45 & 61.27 & 80.10 & 70.01 \\
PROMPTSRC \cite{promptsrc} & 50.71 & 36.59 & 48.53 & 32.53 & 41.35 & 22.46 & 46.84 & 29.32 & 46.86 & 30.23 \\
2LM \cite{2lm} & 51.79 & 27.75 & 42.98 & 28.11 & 41.79 & 12.15 & 50.25 & 28.23 & 46.70 & 24.06 \\
ODG-Net \cite{odg-net} & 42.55 & 32.99 & 49.07 & 21.24 & 49.63 & 17.72 & 45.37 & 31.71 & 46.66 & 25.92 \\
MEDIC \cite{medic} & 48.11 & 30.37 & 46.49 & 27.34 & 39.65 & 15.50 & 45.28 & 27.00 & 44.88 & 25.05 \\
SCI-PD \cite{scipd} & 35.16 & 36.79 & 32.48 & 30.70 & 35.73 & 34.45 & 37.28 & 36.18 & 35.16 & 34.53 \\
ODG-CLIP \cite{odgclip} & \cellcolor{red!20}82.23 & \cellcolor{red!20}87.13 & \cellcolor{red!20}93.46 & \cellcolor{red!20}96.29 & 72.09 & \cellcolor{red!20}81.03 & \cellcolor{red!20}86.80 & \cellcolor{red!20}88.19 & \cellcolor{red!20}83.65 & \cellcolor{red!20}88.16 \\
OSLOPROMPT & \textbf{92.18} & \textbf{94.26} & \textbf{99.60} & \textbf{99.80}& \textbf{85.41} & \textbf{93.50} & \textbf{97.67} & \textbf{92.49} & \textbf{93.72} & \textbf{95.01} \\
\bottomrule
\end{tabular}
\label{table:pacs_domain_performance}
\end{table*}
\section{Potential limitations}
\label{sec:limitations}
We find two potential areas of improvements for \textsc{OSLoPrompt}, as discussed in the following, 
\begin{enumerate}
\item \textbf{Challenges with highly fine-grained datasets:} In fine-grained datasets, where differences between classes are subtle, generating meaningful pseudo-open samples is tricky, and may require more insights in our prompting scheme.

\item \textbf{Impact of pseudo-open image quality:} There is dependence on the Stable diffusion \cite{stablediffusion} model to generate pseudo-open samples. When the prompts are fine-grained and specific, there is a chance that the model can introduce artifacts unrelated to the object in the image.
\end{enumerate}
\clearpage

{
    \small
    \bibliographystyle{ieeenat_fullname}
    \bibliography{main}
}


\end{document}